\documentclass[10pt,twocolumn,letterpaper]{article}

\usepackage{cvpr}              

\usepackage[dvipsnames]{xcolor}

\usepackage{eccvabbrv}

\usepackage{graphicx}
\usepackage{booktabs}
\usepackage{epigraph}
\usepackage{relsize}
\usepackage{mathtools}

\definecolor{cvprblue}{rgb}{0.21,0.49,0.74}
\usepackage[pagebackref,breaklinks,colorlinks,citecolor=cvprblue]{hyperref}

\def\paperID{12} 
\def\confName{3DV\xspace}
\def\confYear{2026\xspace}

\author{Jane Wu\textsuperscript{1} \hspace{1mm} Georgios Pavlakos \textsuperscript{2} \hspace{1mm}Georgia Gkioxari \textsuperscript{3} \hspace{1mm} Jitendra Malik \textsuperscript{1}\\
\textsuperscript{1} UC Berkeley \hspace{1mm}\textsuperscript{2} UT Austin \hspace{1mm} \textsuperscript{3} Caltech
}

\begin{document}


\title{Reconstructing Hand-Held Objects in 3D from Images and Videos}

\twocolumn[{%
\renewcommand\twocolumn[1][]{#1}%
\maketitle
\begin{center}
    \centering
    \captionsetup{type=figure}
    \includegraphics[width=\linewidth]{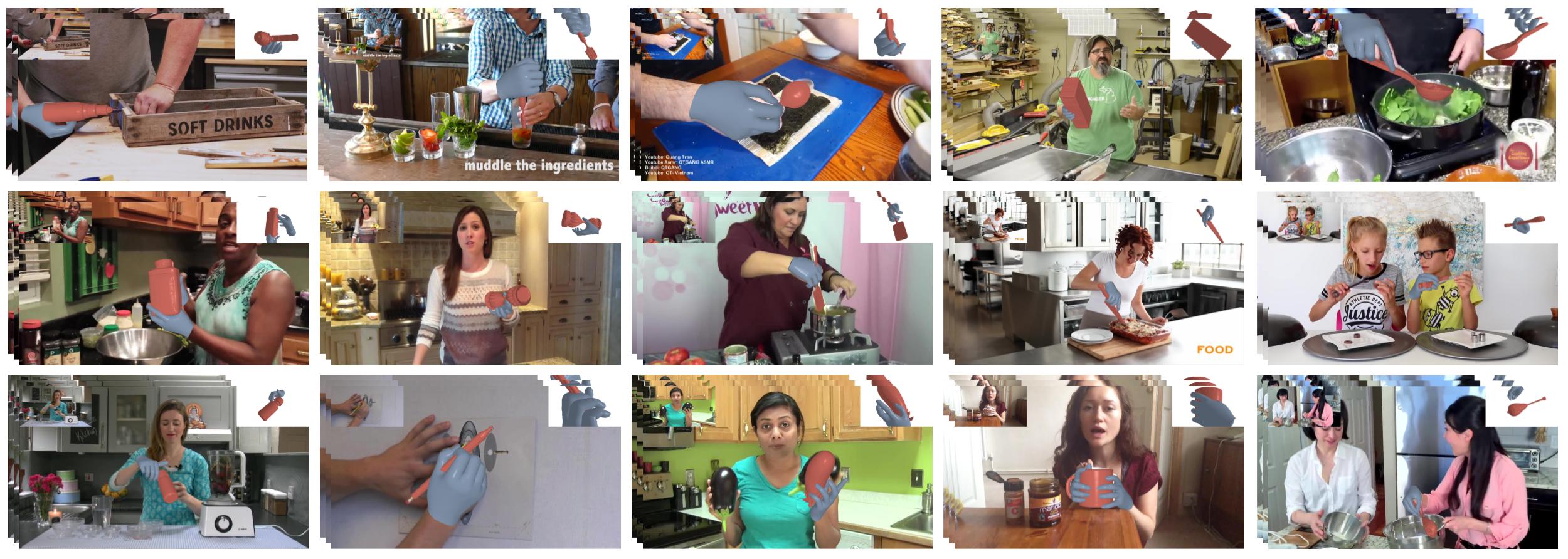}
    \captionof{figure}{We present a scalable approach to hand-held object reconstruction from monocular RGB images or videos that is guided by object recognition and retrieval. Results demonstrate that our method is able to generate realistic object geometry that is also faithful to visual observations and consistent across frames. Please see the supplementary materials for evaluation of temporal consistency.}
    \label{fig:teaser}
\end{center}%
}]

\begin{abstract}
Objects manipulated by the hand (i.e\onedot, manipulanda) are particularly challenging to reconstruct from Internet videos.
Not only does the hand occlude much of the object, but also the object is often only visible in a small number of image pixels.
At the same time, two strong anchors emerge in this setting: (1) estimated 3D hands help disambiguate the location and scale of the object, and (2) the set of manipulanda is small relative to all possible objects.
With these insights in mind, we present a scalable paradigm for hand-held object reconstruction that builds on recent breakthroughs in large language/vision models and 3D object datasets.
Given a monocular RGB video, we aim to reconstruct hand-held object geometry in 3D, over time.
In order to obtain the best performing single frame model, we first present MCC-Hand-Object (MCC-HO), which jointly reconstructs hand and object geometry given a single RGB image and inferred 3D hand as inputs.
Subsequently, we prompt a text-to-3D generative model using a VLM to retrieve a 3D object model that matches the object in the image(s); we call this alignment Retrieval-Augmented Reconstruction (RAR).
RAR provides unified object geometry across all frames, and the result is rigidly aligned with both the input images and 3D MCC-HO observations in a temporally consistent manner.
Experiments demonstrate that our approach achieves state-of-the-art performance on lab and Internet image/video datasets.
We make our
code and models available on the project website: \url{https://janehwu.github.io/mcc-ho}
\end{abstract}

\vspace{-6mm}
\section{Introduction}\label{sec:intro}
Recovering 3D hand-object interactions from visual data is an important problem in both computer vision and robotics.
While there has been significant progress in reconstructing hands~\cite{pavlakos2023reconstructing,potamias2025wilor,ye2025predicting} and objects~\cite{liu2023zero,liu2024one,long2024wonder3d,xu2024instantmesh} separately, joint reconstruction of hand-object interactions presents unique challenges that cannot currently be solved by general reconstruction approaches.
Existing image-to-3D models~\cite{liu2023zero,liu2024one,long2024wonder3d,xu2024instantmesh} are trained on data where the object is usually fully visible in the image; thus, these models are not trained to infer geometry primarily from hand/scene context.
Procedural methods such as Structure from Motion (SfM)~\cite{tomasi1992shape} will also struggle because the object is mostly occluded, and thus it is difficult to obtain enough feature points to achieve accurate reconstruction.
Instead, state-of-the-art approaches in hand-object reconstruction have relied on human assistance for data collection, whether it be lab datasets~\cite{chao2021dexycb,liu2022hoi4d,fan2023arctic} or Internet videos~\cite{cao2021reconstructing,patel2022learning}.
Manually scanned or retrieved object models naturally improve reconstruction quality and can provide better initialization, but this paradigm is not scalable.

In this paper, we present a novel approach to recovering hand-object interactions in 3D from monocular RGB images and videos, focusing in particular on reconstructing hand-held object geometry.
See Figure \ref{fig:teaser} and the supplementary videos.
Given an RGB image and estimated 3D hand, we train a transformer-based model, MCC-Hand-Object (MCC-HO), that jointly infers 3D hand and object geometry.
This geometry is represented as a neural implicit surface composed of occupancy, color, and hand-object segmentation.
By adapting our model architecture from an existing object reconstruction model, MCC \cite{wu2023multiview}, we make use of a learned object prior by fine-tuning an MCC model that is pretrained on the CO3Dv2 dataset \cite{reizenstein2021common}.
Experiments show that our model outperforms existing methods for hand-held object reconstruction on available datasets containing 3D labels, including DexYCB, MOW, and HOI4D \cite{chao2021dexycb,cao2021reconstructing,liu2022hoi4d} (Table \ref{tab:quant_results}, Figures \ref{fig:pointclouds} and \ref{fig:qual_comps}).

Given the limited size of existing 3D hand-object datasets, we additionally leverage vision-language models~\cite{gpt4v,comanici2025gemini} and 3D generative models~\cite{luma,poole2022dreamfusion,chen2024text,long2024wonder3d} in order to improve upon a domain-specific approach.
In conjunction with MCC-HO, we propose Retrieval-Augmented Reconstruction (RAR), a method for automatically retrieving object models using large language/vision models to improve network-inferred geometry.
Specifically, we prompt a VLM~\cite{gpt4v} to recognize the hand-held object in the image and provide a detailed text description.
This description is passed to a text-to-3D generative model~\cite{luma} to obtain a 3D object, which is then rigidly aligned with the input image or video using a combination of DINOv2~\cite{oquab2023dinov2} and inferred 3D cues from MCC-HO.
We demonstrate that RAR leads to quantitative improvements in 3D reconstruction (Tables \ref{tab:video_comparison} and \ref{tab:rar_ablation}), as well as more realistic object geometries compared to existing image-to-3D methods  (Figures \ref{fig:qual_comps} and \ref{fig:genie_results}).

Note that while RAR works well on average, it will fail sometimes if no close enough 3D match is found by the generative model. But the good news is that it does not need to work well on every video for the results of this framework to be useful as a ``data engine'' for robotics. There is a great deal of interest in getting hand-object interaction trajectories for imitation learning by robots \cite{singh2024hand} and typically for this data is collected in lab settings \cite{banerjee2024introducing,fu2025gigahands,hoque2025egodex} which does not result in sufficient diversity. With our paradigm, we can use a quick ``thumbs-up, thumbs-down'' check from a human observer to screen for all videos where the human-object interactions are reconstructed good enough. We have applied this approach with great success to Internet videos from the 100DOH dataset \cite{shan2020understanding}; see the supplementary materials. While these videos will be of slightly lower quality than those in laboratory collected datasets, they have the advantage of scale and diversity and thus can constitute the pre-training phase of a robot data pipeline with fine-tuning provided by the laboratory datasets.

\begin{figure*}[t]
    \centering
    \includegraphics[width=\linewidth]{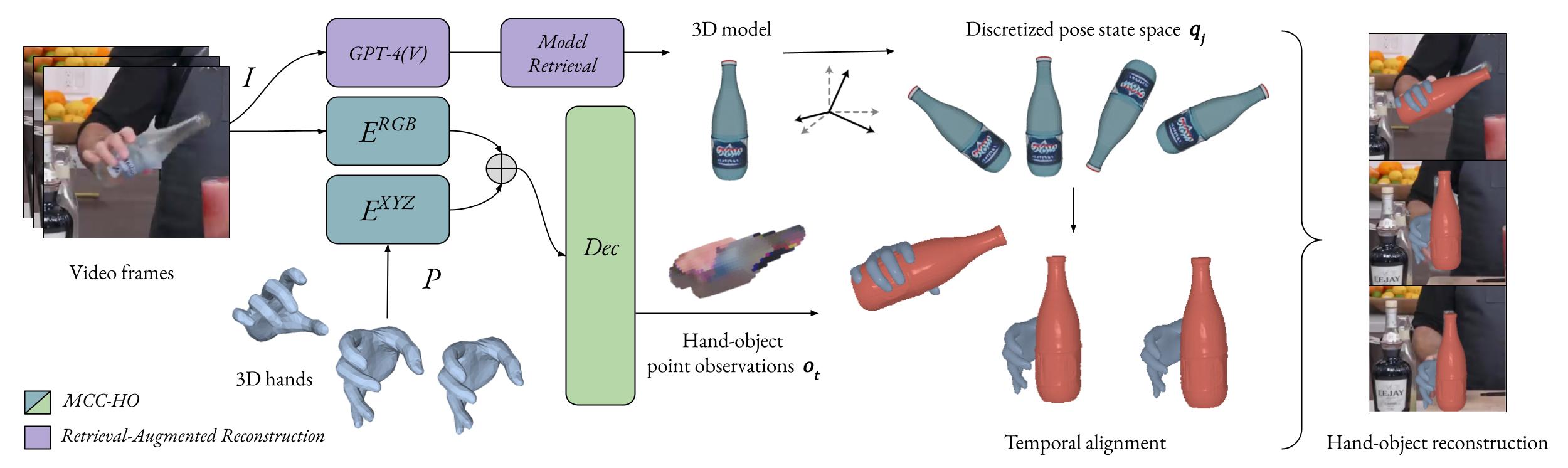}
    \caption{Given an RGB video and estimated 3D hands, our method reconstructs 3D hand-held object trajectories. First, MCC-HO is used to predict hand and object point clouds for each frame (Section \ref{sec:mccho}). Then, a single 3D model for the object is obtained using Retrieval-Augmented Reconstruction (Section \ref{sec:rar}). The 3D object model is rigidly aligned with DINOv2~\cite{oquab2023dinov2} image features and network-inferred geometry in a temporally consistent manner via our Viterbi algorithm (Section \ref{sec:temporal_alignment}).}
    \label{fig:overview}
\end{figure*}

\section{Related Work}\label{sec:related_work}
\textbf{Object reconstruction.}
Reconstructing objects from images is a longstanding problem in computer vision, and the bulk of research in this area has focused on scenarios where the object is fully or almost fully unoccluded in the input view(s).
Structure from motion (SfM) \cite{hartley2003multiple,scharstein2002taxonomy,szeliski2022computer,tomasi1992shape} and SLAM \cite{castellanos1999spmap,mur2015orb,taketomi2017visual} remain robust solutions particularly for multiview reconstruction and camera parameter estimation.
Learning-based object reconstruction methods typically represent 3D geometry as point clouds \cite{qi2017pointnet,qi2017pointnet++}, meshes \cite{gkioxari2019mesh,wen2019pixel2mesh++,worchel2022multi}, voxel grids \cite{xie2019pix2vox,wang2021multi}, or implicit surfaces \cite{park2019deepsdf,mildenhall2021nerf,muller2022instant,tewari2020state,wang2021neus,wu2023multiview}.
Following the release of Objaverse \cite{deitke2023objaverse} and Objaverse-XL \cite{deitke2024objaverse}, a number of works have used this large-scale 3D object dataset to train monocular object reconstruction models \cite{liu2023zero,liu2024one,long2023wonder3d,hong2023lrm}.
The CO3Dv2 dataset \cite{reizenstein21co3d} contains objects from 50 MS-COCO categories \cite{lin2014microsoft} and has also been used to train monocular object reconstruction models \cite{wu2023multiview}.
In practice, we find that off-the-shelf object reconstruction methods do not perform well on images of hand-object interactions due to occlusion from the hand, larger object pose diversity, etc.

\noindent
\textbf{Hand reconstruction.}
Recently, there has been a growing body of literature on estimating hand pose from a monocular RGB image \cite{pavlakos2023reconstructing,li2022interacting,meng20223d,moon2023bringing,wang2023memahand,yu2023acr,zuo2023reconstructing,park2022handoccnet,rong2020frankmocap,prakash20243d} or monocular video \cite{Karunratanakul_2023_CVPR,Wen_2023_CVPR,Xu_2023_CVPR}. These approaches either estimate skeletal joint locations of the hands or use parametric models such as MANO \cite{romero2022embodied}.
Since humans have two hands, other methods aim to reconstruct the left and right hands jointly \cite{Wang_2023_CVPR,Yu_2023_CVPR}.
We use HaMeR~\cite{pavlakos2023reconstructing} in this work as it has achieved state-of-the-art performance on in-the-wild images.

\noindent
\textbf{Model-based hand-object pose estimation.}
When 3D object models are available, the problem of recovering hand-object interactions can be reduced to estimating the pose and shape of the hand and the rigid pose of the object \cite{bhatnagar2022behave,cao2021reconstructing,patel2022learning,fan2023arctic,hasson2020leveraging,hasson2021towards,tekin2019h+,tzionas2016capturing,yang2022artiboost,xie2022chore,tse2022collaborative,qu2023novel}.
The estimated hands can then be used to rigidly align the object with images or videos \cite{cao2021reconstructing,patel2022learning,zhang2023handypriors} via inverse rendering \cite{loper2014opendr,ravi2020accelerating}.
Due to the specific challenges arising from hand-object interactions, e.g.\ highly occluded geometry, optimization-based approaches are considered the current state-of-the-art in this area.
RHO~\cite{cao2021reconstructing} studied the problem of recovering hand-object pose from in-the-wild images, whereas prior work was centered around datasets captured in lab settings.
RHOV~\cite{patel2022learning} extended RHO to video data by adding, among other terms, temporal and optical flow losses.
Aiming to add more physics-inspired constraints, HandyPriors~\cite{zhang2023handypriors} leverages a differentiable contact module to minimize relative sliding between frames.
G-HOP~\cite{ye2024g} guides optimization with a generative hand-object prior.

\noindent
\textbf{Model-free hand-object reconstruction.}
Reconstructing hand-held objects without corresponding 3D models poses additional challenges compared to general object reconstruction due to frequent occlusions by the hand and significantly less publicly available 3D data of hand-object interactions.
Existing approaches are either tailored to reconstruction from a single image \cite{ye2022s,hasson2019learning,chen2022alignsdf,chen2023gsdf,zhang2023moho,fu2023hacd,choi2023handnerf,Prakash2024HOI}, RGB video \cite{huang2022reconstructing,ye2023diffusion,hampali2023hand,ye2024g}, or RGB-D video \cite{wen2023bundlesdf}.
Our approach combines the respective benefits of model-based and model-free reconstruction; we first reconstruct 3D objects from a single image, and the output is used to guide the registration of a retrieved 3D object model.

\noindent
\textbf{Hand-object datasets.}
Datasets containing hand-object interactions are either collected in lab environments \cite{bhatnagar2022behave,hampali2020honnotate,chao2021dexycb,kwon2021h2o,liu2022hoi4d}, through large-scale capture \cite{grauman2022ego4d,grauman2023ego}, or from Internet videos \cite{damen2018scaling,shan2020understanding,cao2021reconstructing}.
A number of hand-held object reconstruction models \cite{chen2022alignsdf,chen2023gsdf,choi2023handnerf} were trained on either the HO-3D \cite{hampali2020honnotate} or DexYCB \cite{chao2021dexycb} datasets.
Since the combined number of object instances in these two datasets is 31, it is difficult for such approaches to generalize to unseen objects.
HOI4D \cite{liu2022hoi4d} is currently the largest available dataset with 3D hand and object labels (also used in \cite{ye2023diffusion}).
There is also a comparably larger amount of data containing videos of hand-object interactions without 3D labels.
Epic Kitchens \cite{damen2018scaling} contains 55 hours of egocentric video of people in their personal kitchens.
Ego4D \cite{grauman2022ego4d} and Ego-Exo4D \cite{grauman2023ego} are also egocentric datasets, and 100 Days of Hands (100DOH) \cite{shan2020understanding} contains 131 days of footage collected from Internet videos.
Following the release of \cite{shan2020understanding}, MOW \cite{cao2021reconstructing} and RHOV \cite{patel2022learning} labeled a subset of images and video clips from 100DOH, respectively, with 3D hands and objects.

\section{Method Overview}
Our approach to reconstructing hand-object interactions from monocular RGB video combines feedforward network inference (MCC-HO) with object model retrieval (RAR) and subsequent rigid alignment.
See Figure \ref{fig:overview}.
First, our transformer-based model, MCC-HO, estimates 3D hand-object geometry from single images (Section \ref{sec:mccho}).
Simultaneously, a 3D object model is automatically synthesized by a text-to-3D generative model~\cite{luma} through a technique we coin Retrieval-Augmented Reconstruction (Section \ref{sec:rar}).
Finally, the 3D object model is rigidly aligned in a temporally consistent manner (Section \ref{sec:temporal_alignment}).

\section{Estimating Hand-Object Geometry}\label{sec:mccho}
In the first stage, hand and object geometry are jointly inferred by training a model adapted from MCC \cite{wu2023multiview} for the specific task of hand-object reconstruction.
Our model, MCC-Hand-Object (MCC-HO), has to deal with two major challenges.
First, MCC assumes RGB-D images as input, whereas our model needs to work with RGB input, since our ultimate goal is to reconstruct from Internet videos.
Second, we have to find a way to cope with the significant domain shift between the training datasets.
MCC was trained on the CO3D dataset~\cite{reizenstein21co3d}, which consists of clean, 360-degree videos where each object is unoccluded and placed on a solid surface.
For MCC-HO, our training data is a relatively small collection of in-the-wild and lab datasets of hand-object interactions, where the hand-held object is at least partially occluded in the vast majority of images.

The inputs to MCC-HO are a single RGB image and 3D hand geometry.
Hand and object segmentation masks can be obtained either from ground truth labels or Segment Anything (SAM/SAM2)~\cite{kirillov2023segment,ravi2024sam}, and the combined hand-object mask determines the bounding box that is used to crop/resize the input image.
Hand geometry is parameterized by the MANO hand model \cite{romero2022embodied}, which is composed of pose parameters $\theta \in \mathbb{R}^{48}$ and shape parameters $\beta \in \mathbb{R}^{10}$. The MANO hand skeleton contains 21 joints.
At training time, the ground truth 3D hand is used.
At test time, HaMeR \cite{pavlakos2023reconstructing} is used to infer the hand geometry.

The inferred geometry is represented as a neural implicit function $\rho(x)$ modeling the hand and object jointly, such that for any 3D location $x$, $\rho$ returns the occupancy probability $\sigma(x)$, RGB color $c(x)$, and segmentation label $m(x)$:
\begin{equation}
    \rho(x) = (\sigma(x), c(x), m(x))
\end{equation}
where $\sigma(x) \in [0, 1]$, $c(x) \in [0, 1]^3$, and $m(x) \in \{0, 1, 2\}$ to indicating a background, hand, or object point.
The scale of the object geometry is normalized with respect to the input 3D hand.

\subsection{Architecture}
MCC-HO has an encoder-decoder architecture that is composed of transformers \cite{vaswani2017attention} and outputs an implicit representation of hand-object geometry.
The input images and 3D hand are passed to the encoder to compute a feature map that is a function of both image and hand embeddings.
These features are then used to condition the decoder, which maps any 3D location to occupancy, color, and hand-object segmentation.
During training, a set of query 3D points $Q$ are uniformly sampled within the bounds of a normalized volume.
At test time, vertices of a voxel grid are passed as query points.

\noindent\textbf{Encoder.} The encoder takes as input a single RGB image $I\in \mathbb{R}^{H\times W \times 3}$ and a posed MANO hand mesh defined in camera world space with vertices $v_h \in \mathbb{R}^{778 \times 3}$ and faces $f_h \in \mathbb{R}^{1538 \times 3}$.
The camera intrinsics are either known or consistent with the estimated 3D hand (\eg, the camera parameters used during HaMeR inference \cite{pavlakos2023reconstructing}).
In order to jointly encode the image and hand geometry, the hand mesh is rasterized and per-pixel 3D points for the hand are sampled.
For each pixel $(i, j)$ in the image, a ray is cast from the camera aperture to the pixel center to determine the first intersection point with the hand mesh at a point $P_0(i,j)$ in camera world space.
If an intersection exists, the 3D location of each point is barycentrically interpolated from the intersecting camera world space triangle vertices, denoted $\{v_{h,1}, v_{h,2}, v_{h,3}\}$, such that:
\begin{equation}
    P_0(i,j) = \alpha_1 v_{h,1}+ \alpha_2 v_{h,2} + \alpha_3v_{h,3} \in \mathbb{R}^3,
\end{equation}
with barycentric weights $\alpha$.
The points $P_0$ are used to define a normalized camera world space, such that the 3D coordinates passed to the network are $P = s(P_0 - \mu(P_0)) / \sigma(P_0)$ and geometry is inferred in this same coordinate system.
Importantly, $s$ is a constant scaling factor used to ensure the object geometry can be contained by the bounds of this hand-normalized sampling space.

The image $I$ and hand points $P$ are encoded into a single representation $R$, where:
\begin{equation}
    R \coloneqq f(E^{RGB}(I), E^{XYZ}(P)) \in \mathbb{R}^{N^{enc} \times C},
\end{equation}
and $E^{RGB}$ and $E^{XYZ}$ are identical to the transformers proposed in MCC \cite{wu2023multiview}, though we define $P$ differently, \ie, without an input depth image.
The image transformer $E^{RGB}$ has a Vision Transformer (ViT) architecture \cite{dosovitskiy2020image} with $16\times 16$ patch embeddings.
The point transformer $E^{XYZ}$ uses a self-attention-based patch embedding design that differentiates between seen and unseen pixel points.
In particular, for points in $P$ where no ray-mesh intersection was found, a special $C$-dimensional embedding is learned in lieu of embedding 3D points.
The function $f$ concatenates the output of the two transformers, each of which has dimensions $N^{enc} \times C$.
$N^{enc}$ is the number of tokens used in the transformers and $C$ is the output channel size.

\noindent\textbf{Decoder.} The decoder takes as input the encoder output, $R$, as well as a set of query points $Q \in \mathbb{R}^{N_q \times 3}$ (we choose $N_q = 1024$).
These inputs are passed to a transformer with architecture adopted from MCC and inspired by Masked Autoencoders (MAE) \cite{he2022masked}.
Each query token is passed to three output heads that infer binary occupancy $\sigma$, RGB color $c$, and hand-object segmentation $m$, such that:
\begin{equation}
    Decoder(R, Q) \coloneqq (\sigma(Q), c(Q), m(Q)).
\end{equation}
The decoder layers that output occupancy and color align with MCC, but MCC-HO has an additional linear layer that infers multiclass segmentation of each point into either background, hand, or object labels.

As discussed in further detail in MCC, a point-query-based decoder presents a number of advantages with respect to computational efficiency and scalability.
Since the encoder is independent of the point queries, any number of points can be sampled using the same encoder structure and feature map size.
The decoder architecture is relatively lightweight and also enables dynamic changes to the point sampling resolution and query size at training and/or test time.

\noindent\textbf{Losses.}
The training loss is a combination of occupancy, color, and segmentation:
\begin{equation}
    \mathcal{L} = \mathcal{L}_{\sigma} + \mathcal{L}_{c}(\sigma_{gt}) + \mathcal{L}_{m}
\end{equation}
The occupancy loss $\mathcal{L}_{\sigma}$ is a binary cross-entropy loss comparing the predicted and ground truth occupancies of each query point.
The color loss $\mathcal{L}_c$ is a 256-way classification loss computed for query points that have ground truth occupancy of 1.
The segmentation loss $\mathcal{L}_m$ is 3-way classification loss for each query point corresponding to background (empty space), hand, and object labels.

\section{Retrieval-Augmented Reconstruction}\label{sec:rar}
Like any other deep learning model, how well MCC-HO is expected to perform is related to the amount of training data available, \eg, the size of available hand-object datasets.
These are currently all quite small, orders of magnitude smaller than the ``big data'' on which computer vision foundation models have been trained.
How might we find a way to exploit these other models to our end?
Specifically, recognition of objects is very well developed~\cite{gpt4v}, and large collections of 3D object models~\cite{deitke2023objaverse,deitke2024objaverse} exist and have facilitated advancements in 3D generative modeling~\cite{luma,poole2022dreamfusion,chen2024text,long2024wonder3d}.
Thus, in order to further improve upon network-inferred object geometry, we exploit these ``big data'' models by using GPT-4(V)~\cite{gpt4v} and a text-to-3D generative model~\cite{luma} to automatically detect and ``retrieve'' a 3D object corresponding to the hand-held object in an image.
We call this technique Retrieval-Augmented Reconstruction (RAR), somewhat in analogy to Retrieval-Augmented Generation in natural language processing \cite{lewis2020retrieval}.

Given a representative frame of the video, we prompt GPT-4(V) to provide a detailed description of the object the hand is holding.
This text caption is passed to Genie, a text-to-3D model from Luma \cite{luma}, to retrieve realistic 3D object geometry and appearance (Genie is also used in~\cite{li2023object}).
Since Genie outputs geometry normalized to fit inside a unit cube, the scale of the 3D model must be adjusted with respect to the input 3D hand.

\subsection{Rigid Alignment}\label{sec:rigid_alignment}
After a 3D object model is retrieved, it is rigidly aligned with the input visual data and MCC-HO predictions.
We begin by considering the single image case.
First, the object scale is estimated using the predicted MCC-HO object point cloud, which is a reasonable approximation because the prediction is grounded by the input 3D hand. 
If the MCC-HO point cloud is $X \in \mathbb{R}^{N\times 3}$ and a sampled point cloud of the retrieved object model is $Y \in \mathbb{R}^{N\times 3}$, then the scale is defined as $s = \sqrt{\lambda_{X}^0} / \sqrt{\lambda_{Y}^0}$
where $\lambda_{X}^0$ is the largest eigenvalue of $X^TX$ and $\lambda_{Y}^0$ is similarly defined for $Y$.

The remaining parameters to be optimized are object rotation $R$ and translation $T$.
In order to (1) guarantee a globally optimal solution while (2) maintaining fixed compute time, we discretize the space of all possible parameter values to construct finite state spaces $\{\mathcal{R}, \mathcal{T}\}$.
The 3D rotation group, SO(3), is discretized using the unit cube subdivision strategy proposed in \cite{kurz2017discretization}.
Translation is defined as an offset from the mean location of the MCC-HO point cloud $\mu(X)$, and the space is discretized by constructing a small voxel grid centered at $\mu(X)$.
In order to avoid intractably large state space sizes (\eg, $\mathbb{R}^3 \times \mathbb{R}^3$), we first optimize for rotation and then translation.
For each parameter, we select the state that results in the lowest error using the sum of the following two metrics.

\noindent\textbf{MCC-HO Chamfer Distance.} The Chamfer Distance (CD) between the MCC-HO point cloud and a transformed object model is the usual 3D evaluation metric.
Note that when optimizing for rotation, the object model (centered at the origin) is first rotated and then translated to the mean location of the MCC-HO point cloud.

\begin{figure}[t]
    \centering
    \includegraphics[width=\linewidth]{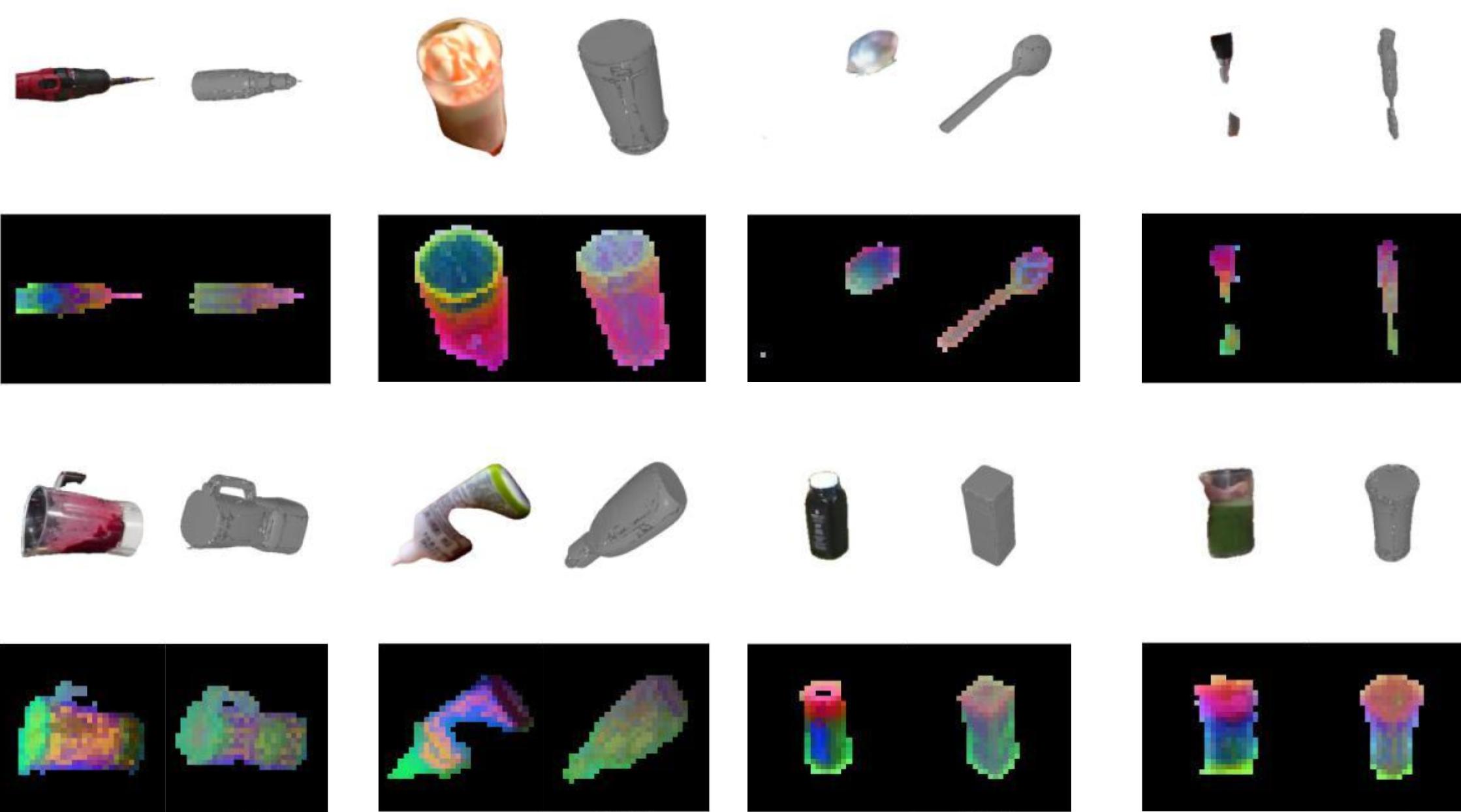}
    \caption{We compute a PCA basis of DINOv2 features using all frames masked by the object silhouettes (one frame and its first three PCA components are shown on the left side of each example). The first three components of this basis are used to determine the maximum likelihood Genie object rotation for each frame (shown on the right side).}
    \label{fig:rar}
    \vspace{-3mm}
\end{figure}

\noindent\textbf{DINOv2 PCA Similarity.}
The MCC-HO point cloud gives us a reasonable estimate for object pose, but using this signal alone may not result in poses that align well with the input image (particularly for geometry with symmetries and in-the-wild objects).
Thus, we additionally use DINOv2~\cite{oquab2023dinov2} features to determine visual similarities between a transformed object model and the input image.
For each rotation/translation state, the transformed object mesh is rendered using Pytorch3D~\cite{ravi2020accelerating}.
A PCA basis is constructed using the DINOv2 features of the input image, and the first three PCA components are computed for both the rendered features $\mathcal{F}_j$ and input image features $\mathcal{F}_0$.
We compute the cosine similarity between the two as
\begin{equation}
 {\textstyle
    E_{DINO} = 1-\frac12\left(\frac{\text{PCA}(\mathcal{F}_j)\cdot \text{PCA}(\mathcal{F}_0)}{\max(||\text{PCA}(\mathcal{F}_j)||_2 \cdot ||\text{PCA}(\mathcal{F}_0)||_2), \epsilon)} + 1\right) }
\end{equation}
where $\mathcal{F}_j$ and $\mathcal{F}_0$ are both masked by the input image object silhouette.
When ground truth object masks are not available, we use SAM 2~\cite{ravi2024sam}.
See Figure~\ref{fig:rar} for rotation results using example images from 100DOH~\cite{shan2020understanding}.

\section{From Images to Videos}\label{sec:temporal_alignment}
In the previous section, we showed how given a single image, we select the rotation and translation of the 3D object model that minimizes error metrics based on CD and DINOv2 similarity.
Now, if we have as input a video stream, we can further exploit the additional cue of temporal smoothness---we expect adjacent frames to have rotation and translation states that are close to each other. 

Given a sequence of video frames with timesteps $t$, we jointly optimize for per-frame rotations $R(t)$ and subsequently per-frame translations $T(t)$.
The typical approach for temporal alignment is to use gradient-based optimization ~\cite{patel2022learning,ye2023diffusion}, which is prone to issues with local minima and has unbounded computation time.
Aiming for robustness and efficiency, we propose using the Viterbi algorithm~\cite{viterbi1967error,forney1973viterbi,jurafsky2000speech} to solve for a global solution to rigid object alignment across all frames of a video sequence.
The Viterbi algorithm is a dynamic programming approach to obtaining the maximum \textit{a posteriori} probability of the most likely sequence of hidden states $Q = q_1, q_2, \ldots, q_N$ that results in a sequence of observed events $O = o_1, o_2, \ldots, o_N$.
The hidden states $q_t$ can take on values from a finite set of parameters $\mathcal{Q}$ (where in our case, $\mathcal{Q}=\mathcal{R}$ or $\mathcal{Q}=\mathcal{T}$).
Following the single image case, the observations $o_t$ are a combination of the MCC-HO inferred object point clouds and the DINOv2 image features.

Each cell of the Viterbi trellis, $v_t(j)$, is the probability of ending up in state $j$ after seeing the first $t$ observations and passing through the most probable state sequence $q_1,\ldots,q_{t-1}$.
Given the probability of being in every state at time $t-1$, the Viterbi probabilities are computed via:
\begin{equation}
    v_t(j) = \max_{i=1}^{|\mathcal{Q}|} \hspace{1mm} v_{t-1}(i)a_{ij}b_j(o_t),
\end{equation}
where $a_{ij}$ is the transition probability from the previous state $q_i$ to the current state $q_j$, and $b_j(o_t)$ is the state observation likelihood given the current state~$j$.
In particular, $a_{ij}$ is Rodrigues' rotation error when optimizing for rotations and Euclidean distance when optimizing for translations.
The emission cost $b_j(o_t)$ is the sum of the two metrics defined in the single image discussion, \eg, CD with respect to the MCC-HO point clouds and DINOv2 PCA similarity with respect to the input images.

\section{Experiments}
Section \ref{sec:datasets} details the datasets that were used for training and evaluation.
Section \ref{sec:metrics} defines the evaluation metrics used for comparisons.
We also quantitatively (Section \ref{sec:quant}) and qualitatively (Section \ref{sec:qual}) evaluate against existing work and find that our approach achieves state-of-the-art performance on all three datasets used \cite{chao2021dexycb,cao2021reconstructing,liu2022hoi4d}.
Also in Section \ref{sec:rar_results}, we provide qualitative results for our Retrieval-Augmented Reconstruction (RAR) approach applied to in-the-wild Internet videos from 100DOH~\cite{shan2020understanding} to demonstrate the feasibility and scalability of our automated model retrieval and fitting.
Section \ref{sec:ablations} includes ablation studies for MCC-HO, improvements attributed to RAR, and the effect of conditioning hand-held object reconstruction on 3D hands.

\begin{table*}[t]
    \footnotesize
    \begin{center}
    \begin{tabular}{ccccc|cccc|cccc}
    \hline
    \multicolumn{1}{c}{} & \multicolumn{4}{c}{DexYCB} & \multicolumn{4}{c}{MOW} & \multicolumn{4}{c}{HOI4D} \\
     & F-5 ($\uparrow$) & F-10 ($\uparrow$) & CD ($\downarrow$) & Vol ($\downarrow$) & F-5 ($\uparrow$) & F-10 ($\uparrow$) & CD ($\downarrow$)& Vol ($\downarrow$) & F-5 ($\uparrow$) & F-10 ($\uparrow$) & CD ($\downarrow$) & Vol ($\downarrow$) \\
    \hline
    HO~\cite{hasson2019learning} & 0.24 & 0.48 & 4.76 & 11.8 & 0.03& 0.06 & 49.8 & 78.2 & 0.28 & 0.51 & 3.86 & 12.7\\
    IHOI~\cite{ye2022s} & - & - & -  & - & 0.13 & 0.24 & 23.1 & 20.4 & 0.42 & 0.70 & 2.7 & \textbf{0.91} \\
    HORSE~\cite{Prakash2024HOI} & 0.23 & 0.42 & 6.97  & 3.63 & 0.11 & 0.23 & 24.5 & 17.0 & 0.26 & 0.45 & 6.69 & 1.39\\
\textbf{MCC-HO (Ours)} & \textbf{0.36} & \textbf{0.60} &  \textbf{3.74} & \textbf{2.42}&  \textbf{0.15} & \textbf{0.31} & \textbf{15.2} & \textbf{13.8} & \textbf{0.52} & \textbf{0.78} & \textbf{1.36} & 1.94\\
    \end{tabular}
    \end{center}
    \vspace{-5mm}
    \caption{We compare our method, MCC-HO, to prior works on held-out test images from DexYCB, MOW, and HOI4D. Chamfer Distance (cm$^2$), F-score (5mm, 10mm), and Intersection Volume using 0.5cm voxels (as in HO~\cite{hasson2019learning}, IHOI~\cite{ye2022s}) are reported.}
    \label{tab:quant_results}
\end{table*}

\begin{table}[t]
\small
    \begin{center}
    \begin{tabular}{c|ccc}
    \hline
    Method & F-5 ($\uparrow$)& F-10 ($\uparrow$)& CD ($\downarrow$)\\
    \hline
    G-HOP~\cite{ye2024g} & 0.61 & 0.89 & 0.8 \\
    \textbf{MCC-HO + RAR} & \textbf{0.74} & \textbf{0.91} & \textbf{0.64} \\
    \end{tabular}
    \end{center}
    \vspace{-5mm}
    \caption{Quantitative comparison to a state-of-the-art video-based hand-held object reconstruction method using the HOI4D test dataset. Mean metrics are reported for consistency with G-HOP.}
    \vspace{-4mm}
    \label{tab:video_comparison}
\end{table}

\subsection{Datasets}\label{sec:datasets}
We experiment on four different datasets for hand-object reconstruction.

\noindent\textbf{HOI4D.} The HOI4D dataset \cite{liu2022hoi4d} consists of 4,000 lab videos of hands interacting with specified categories of common objects. We selected all sequences containing one of 6 hand-held object categories, and divide the sequences into training, validation and test sets.
In total, we used 1142 training sequences, 6 validation sequences, and 12 test sequences consistent with DiffHOI \cite{ye2023diffusion}.
For the validation set, we selected one video per category and favored object instances that appear the least frequently in the dataset.

\noindent\textbf{DexYCB.} The DexYCB dataset \cite{chao2021dexycb} contains 1,000 lab videos of hands interacting with 21 YCB objects. We used the ``s0'' dataset split, which contains 800 training, 40 validation, and 160 test sequences.

\noindent\textbf{MOW/RHOV.} MOW \cite{cao2021reconstructing} is a dataset of 512 images from 100 Days of Hands \cite{shan2020understanding} labeled with 3D hands and objects.
We use the same splits as \cite{ye2022s} with 350 training images and 92 test images.
Additionally, we use 110 videos from RHOV \cite{patel2022learning}, each of which is a 2 second clip centered at the corresponding MOW frame.
In the RHOV dataset, 19 videos overlap with the MOW test set and thus we do not use them during training.

\noindent\textbf{100DOH.} The 100 Days of Hands dataset \cite{shan2020understanding} contains 131 days of footage collected from Internet videos showing hands interacting with objects. The vast majority of videos do not have associated 3D hand or object labels, so we evaluate how well our approach generalizes using examples from this unlabeled subset.

For DexYCB \cite{chao2021dexycb}, MOW \cite{cao2021reconstructing}, and HOI4D \cite{liu2022hoi4d}, all results generated using our approach use the ground truth 3D hand provided by each dataset.
For 100DOH \cite{shan2020understanding} results (\eg, examples not contained in MOW), we use HaMeR \cite{pavlakos2023reconstructing} to estimate the 3D hand.

\subsection{Evaluation Metrics}\label{sec:metrics}
Since the focus of our approach is on the object manipulated by the hand, we quantitatively evaluate object reconstruction quality using standard 3D metrics: CD (cm$^2$) and F-score reported at 5mm and 10mm.
Both metrics are computed on point sets.
The 3D hand provided by each dataset is used as input; note that MOW uses FrankMocap~\cite{rong2020frankmocap} and HOI4D optimizes hand pose using 2D keypoint annotations.
For each example, we sample 10k points on the surface of the ground truth mesh consistent with~\cite{ye2022s,ye2023diffusion} (no transformations are applied to the ground truth geometry).
The MCC-HO predicted point clouds are supersampled/subsampled to obtain 10k points as well.
In order to account for unknown ground truth camera parameters, ICP with scaling~\cite{ravi2020accelerating} is used to align all predicted geometry with the ground truth mesh.
Median metrics are reported unless otherwise noted.

\subsection{Quantitative Evaluation}\label{sec:quant}
We quantitatively evaluate MCC-HO on a variety of labeled hand-object datasets and in comparison to existing model-free reconstruction approaches \cite{hasson2019learning,ye2022s,Prakash2024HOI}.
See Table \ref{tab:quant_results}.
In order to sample points more densely in hand-object regions, the dimensions and granularity of the voxel grid used to query MCC-HO are determined from an initial coarse prediction using a default voxel grid.
For DexYCB and HOI4D, we use every 5th frame of each test video sequence.
The results for IHOI \cite{ye2022s} and HO \cite{hasson2019learning} evaluated on the MOW dataset are obtained from the IHOI and DiffHOI \cite{ye2023diffusion} papers; otherwise, we used available public repositories of prior work.
Our method achieves state-of-the-art performance for the Chamfer Distance and F-score metrics on all three datasets, implying that our technique is not overfitted to any particular dataset.
We conjecture that increasing the number of unique object instances seen during training is important to model generalization.
Thus, the pretrained MCC model provides a strong object prior for initialization.

We also compare to G-HOP~\cite{ye2024g}, a state-of-the-art video-based approach for RGB hand-held object reconstruction in Table \ref{tab:video_comparison}.
Our method (using generated Genie objects for RAR) outperforms the generalist G-HOP model (that was trained on seven hand-object datasets~\cite{brahmbhatt2020contactpose,chao2021dexycb,corona2020ganhand,liu2022hoi4d,taheri2020grab,yang2022oakink}) in all three metrics.

\subsection{Qualitative Evaluation} \label{sec:qual}
\begin{figure}[t]
    \centering
    \begin{subfigure}[b]{0.3\linewidth}
    \includegraphics[width=\linewidth]{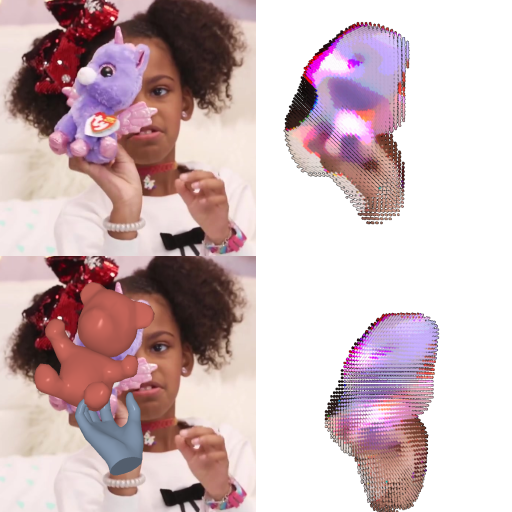}
    \end{subfigure}
    \hfill
    \begin{subfigure}[b]{0.3\linewidth}
    \includegraphics[width=\linewidth]{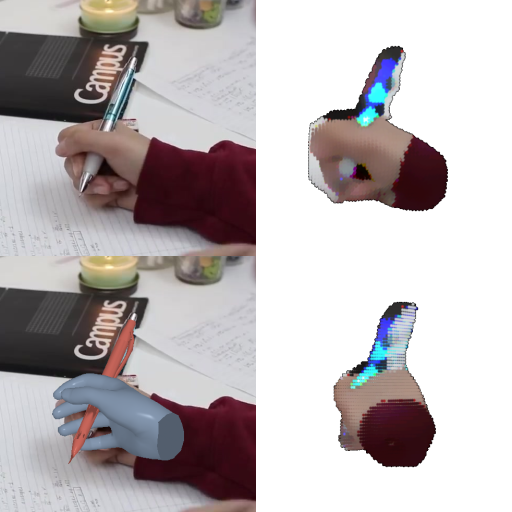}
    \end{subfigure}
    \hfill
    \begin{subfigure}[b]{0.3\linewidth}
    \includegraphics[width=\linewidth]{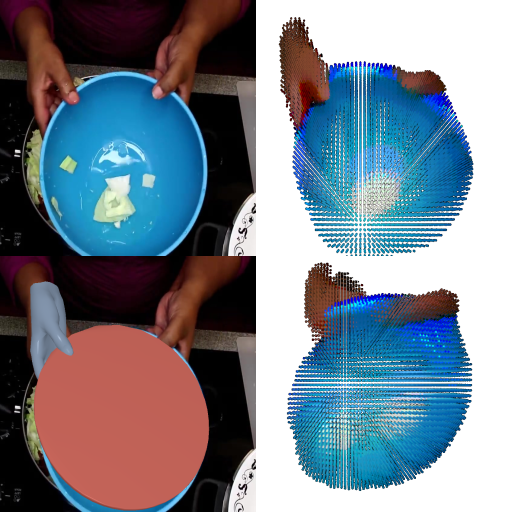}
    \end{subfigure}
    \hfill
    \caption{MCC-HO results for MOW test examples. The input image (top, left), network-inferred hand-object point cloud (top, right), RAR (bottom, left), and an alternative view of the point cloud (bottom, right) are shown.}
    \label{fig:pointclouds}
\end{figure}

\begin{figure}[t]
\vspace{-1mm}
    \centering
    \includegraphics[width=\linewidth]{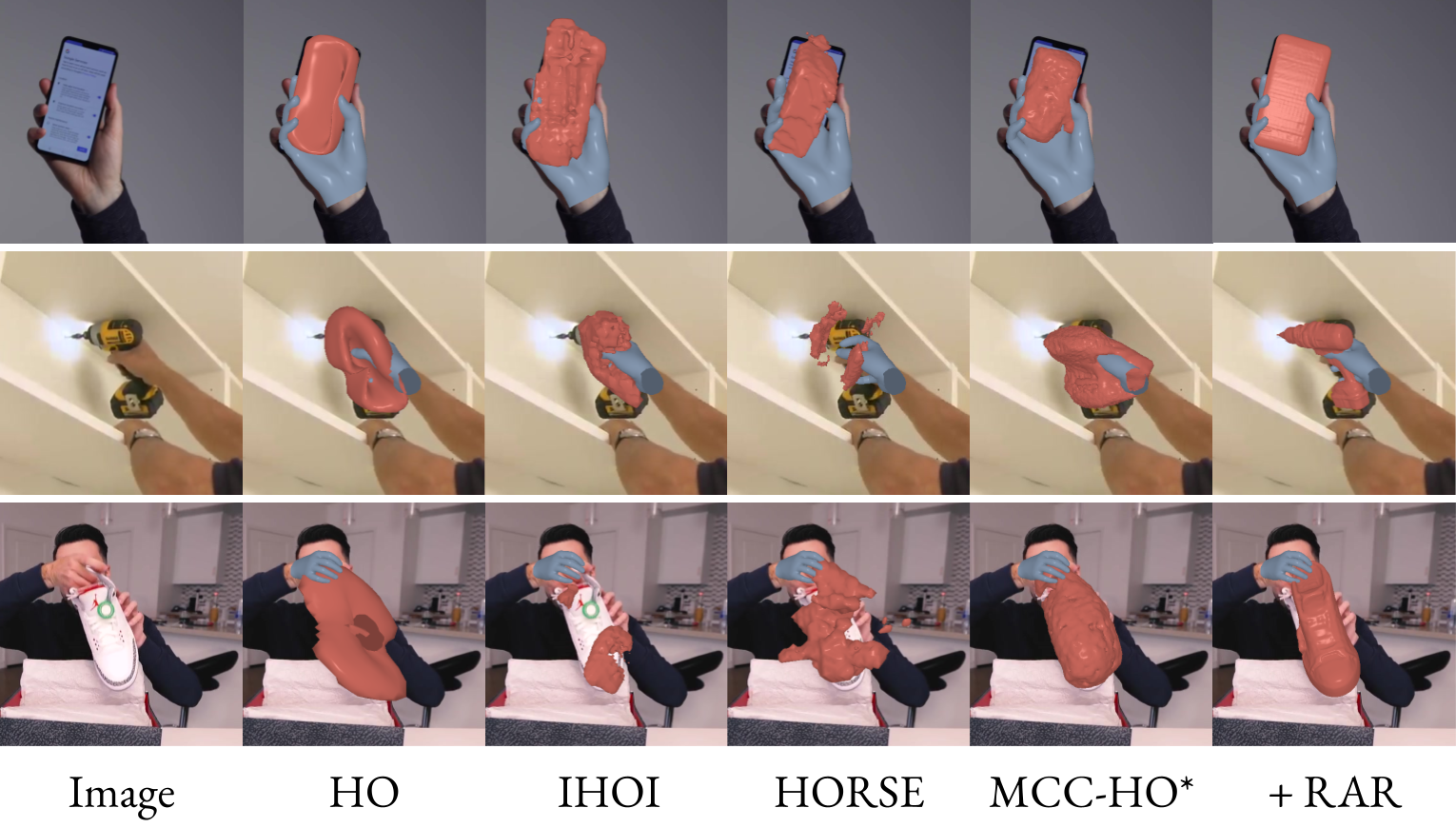}
    \hfill
    \vspace{-5mm}
    \caption{Qualitative comparisons using MOW test examples. *Note that the MCC-HO point clouds are rendered as a mesh via Poisson surface reconstruction, which can introduce artifacts not attributed to our method. The last column is MCC-HO + RAR.
    }
    \label{fig:qual_comps}
    \vspace{-5mm}
\end{figure}

We qualitatively evaluate (1) the reconstruction quality of MCC-HO and (2) how well retrieved object models using RAR can be rigidly aligned to the predicted point clouds in 3D (a necessary and sufficient condition for downstream tasks).
Figure \ref{fig:pointclouds} shows predicted point clouds for a number of MOW test examples, as well as the result of RAR.
Since MCC-HO is trained with monocular images, the predicted RGB colors are only valid when viewed from the given camera.
Because the object geometry is conditioned on the 3D hand, our approach tends to predict reasonable hand-object contact.
Additionally, the object in contact with the hand is correctly identified by MCC-HO without input segmentation maps.
Figure \ref{fig:qual_comps} shows qualitative comparisons to existing work, where the combination of MCC-HO + RAR allows us to obtain more realistic object geometry.

\begin{figure*}[!ht]
    \centering
    \begin{subfigure}[b]{0.31\linewidth}
    \centering
    \includegraphics[width=\linewidth]{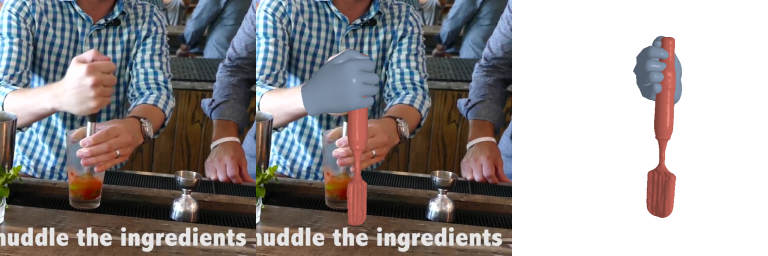}\\
    \vspace{1mm}
    \includegraphics[width=\linewidth]{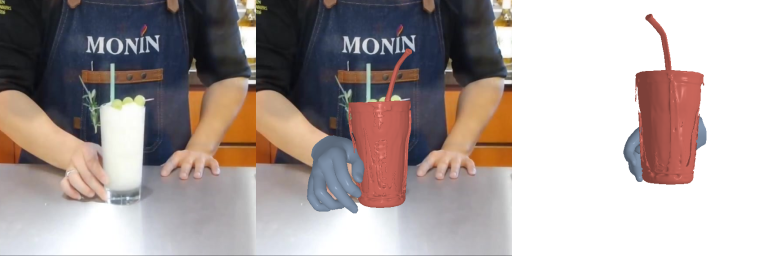}\\
    \vspace{1mm}
    \includegraphics[width=\linewidth]{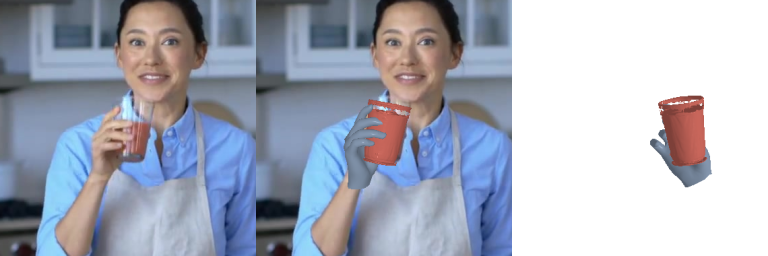}
    \end{subfigure}
    \hfill
    \begin{subfigure}[b]{0.31\linewidth}
    \centering
    \includegraphics[width=\linewidth]{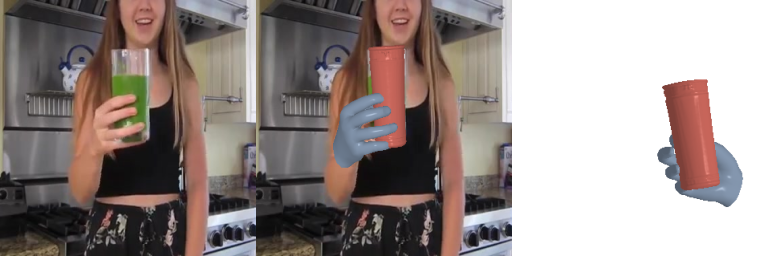}\\
    \vspace{1mm}
    \includegraphics[width=\linewidth]{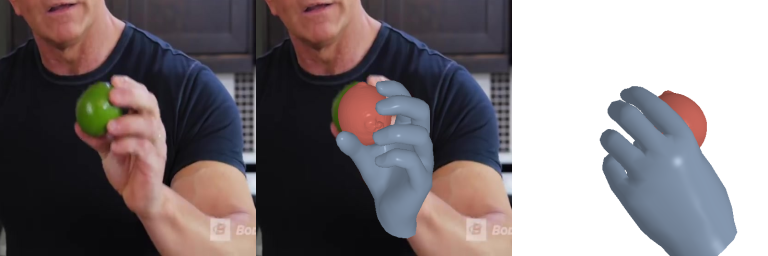}\\
    \vspace{1mm}
    \includegraphics[width=\linewidth]{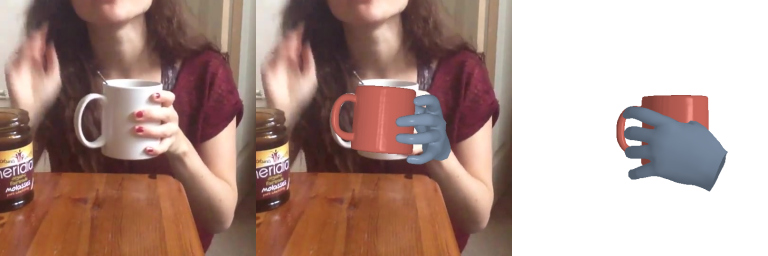}
    \end{subfigure}
    \hfill
    \begin{subfigure}[b]{0.31\linewidth}
    \centering
    \includegraphics[width=\linewidth]{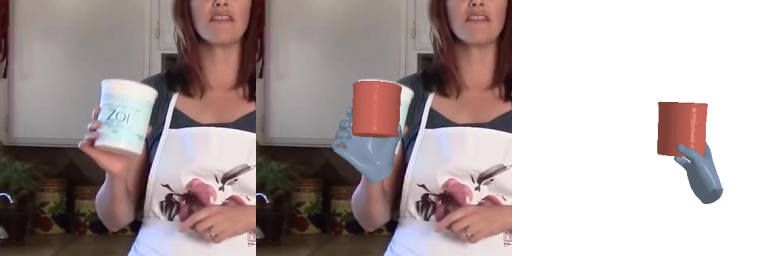}\\
    \vspace{1mm}
    \includegraphics[width=\linewidth]{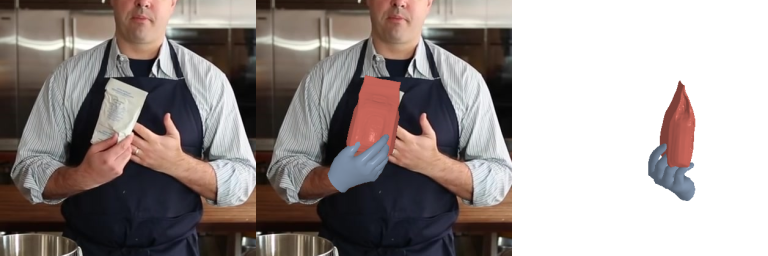}\\
    \vspace{1mm}
    \includegraphics[width=\linewidth]{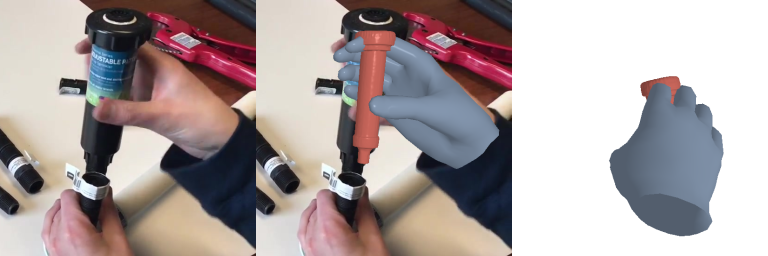}
    \end{subfigure}
    \caption{Our approach applied to in-the-wild images from the 100 Days of Hands dataset \cite{shan2020understanding}. For each image, we estimate 3D hands using HaMeR \cite{pavlakos2023reconstructing}, inference object geometry using MCC-HO and retrieve a 3D model using RAR (\eg, GPT-4(V) and Genie).}
    \label{fig:genie_results}
\end{figure*}

\subsubsection{Retrieval-Augmented Reconstruction}\label{sec:rar_results}
The ultimate goal of RAR is to provide a scalable paradigm for creating large hand-object 3D datasets; to this end, we explore the potential to scale up our approach on unlabeled images and videos.
Figure \ref{fig:genie_results} shows our results for 9 images from 100DOH \cite{shan2020understanding} that do not have 3D ground truth (\eg, not in MOW).
Estimated 3D hands for each image are obtained using HaMeR \cite{pavlakos2023reconstructing}.
At test time, our method does not require an input hand-object segmentation mask \cite{cao2021reconstructing,patel2022learning}, depth image \cite{wu2023multiview}, or hand joint locations \cite{ye2022s,choi2023handnerf}.
Please see the supplementary materials for additional results.

\subsection{Ablation Studies}\label{sec:ablations}

\noindent\textbf{MCC Baseline.}
Because our model is built on top of MCC \cite{wu2023multiview}, we begin by evaluating whether MCC-HO leads to performance gains for the task of hand-held object reconstruction.
For MCC, we use the released model that was trained on the CO3Dv2 dataset \cite{reizenstein21co3d} and apply an existing monocular depth estimator \cite {Ranftl2021} to obtain depth maps (as MCC takes an RGB-D image as input).
Table \ref{tab:mcc_ablation} compares 3D metrics between MCC and MCC-HO computed on the MOW test set.
We find that using MCC-HO, which was trained on the specific task of hand-held object reconstruction, leads to performance improvements on test images.

\begin{table}[t]
\small
    \begin{center}
    \vspace{-3mm}
    \begin{tabular}{c|ccc}
    \hline
     Method & F-5 ($\uparrow$)& F-10 ($\uparrow$)& CD ($\downarrow$)\\
    \hline
    MCC \cite{wu2023multiview} & 0.10 & 0.21 & 31.6  \\
    \textbf{MCC-HO} & \textbf{0.15} & \textbf{0.31} & \textbf{15.2}\\
    \end{tabular}
    \end{center}
    \vspace{-3mm}
    \caption{Analysis of MCC-HO performance gains over our object reconstruction baseline, MCC, using the MOW test dataset.}
    \label{tab:mcc_ablation}
    \vspace{-2mm}
\end{table}

\noindent\textbf{Effectiveness of Input Hand.}
Table~\ref{tab:hand_ablation} analyzes the effectiveness of hand pose prediction input by evaluating MCC-HO with and without the input hand mesh.
As we expect, the model performs significantly worse without the input hand geometry.
The experimental results further corroborate the hypothesis that grounding the 3D object prediction with respect to the hand coordinate system significantly constrains the reconstruction problem.

\begin{table}[t]
\small
    \begin{center}
    \begin{tabular}{c|ccc}
    \hline
     Model Parameters & F-5 ($\uparrow$)& F-10 ($\uparrow$)& CD ($\downarrow$)\\
    \hline
    MCC~\cite{wu2023multiview} & 0.10 & 0.21 & 31.6  \\
    \textbf{MCC-HO (w/o hand)} & 0.03 & 0.12 & 71.6  \\
    \textbf{MCC-HO} & \textbf{0.15} & \textbf{0.31} & \textbf{15.2}\\
    \end{tabular}
    \end{center}
    \vspace{-6mm}
    \caption{Analysis of how hand-held object reconstruction benefits from conditioning on the hand using the MOW test dataset.}
    \label{tab:hand_ablation}
    \vspace{-5mm}
\end{table}

\noindent\textbf{RAR Reconstruction Improvements.}
RAR leverages text-to-3D generative models to synthesize realistic geometry, and we quantitatively evaluate whether using RAR in conjunction with MCC-HO leads to improved results. Table \ref{tab:rar_ablation} demonstrates that RAR (with Genie objects) can indeed help improve the object reconstruction quality of monocular reconstruction methods such as MCC-HO.

\begin{table}[t]
\vspace{-2mm}
\small
    \begin{center}
    \begin{tabular}{c|ccc}
    \hline
    Method & F-5 ($\uparrow$)& F-10 ($\uparrow$)& CD ($\downarrow$)\\
    \hline
    MCC-HO & 0.15 & 0.31 & 15.2\\
    MCC-HO + RAR & \textbf{0.16} & \textbf{0.44} & \textbf{9.41}\\
    \end{tabular}
    \end{center}
    \vspace{-5mm}
    \caption{Analysis of how RAR improves object reconstruction. Median Chamfer Distance (cm$^2$) and F-score (5mm, 10mm) metrics are computed on the MOW test dataset.}
    \label{tab:rar_ablation}
    \vspace{-5mm}
\end{table}

\section{Conclusion}
In this paper, we present a novel paradigm for reconstructing hand-held objects that combines the respective benefits of model-free and model-based prediction.
Our transformer-based model, MCC-HO, is trained to predict hand-object geometry given a single RGB image and estimated 3D hand.
Experiments conducted using the DexYCB, MOW, and HOI4D datasets all demonstrate that MCC-HO achieves state-of-the-art performance in hand-held object reconstruction.
We additionally present Retrieval-Augmented Reconstruction (RAR), an automatic method for object model retrieval that leverages recent advances in large language/vision models and 3D object datasets.
These two approaches can be combined to scale the amount of labeled hand-object interaction data, as suggested by our results using unlabeled 100DOH videos~\cite{shan2020understanding}.

\noindent\textbf{Limitations.}
As 3D generative models improve, RAR will naturally be able to retrieve more accurate and realistic geometry.
Our approach also relies on the ability to obtain accurate hand and object masks which are tracked across frames.
We expect that both MCC-HO and our rigid alignment algorithm will reap the benefits of any future image segmentation/tracking improvements.

\section*{Acknowledgements}
We would like to thank Ilija Radosavovic for helpful discussions and feedback.
This work was supported by ONR MURI N00014-21-1-2801.
J.\ W.\ was supported by the NSF Mathematical Sciences Postdoctoral Fellowship and the UC President’s Postdoctoral Fellowship.

{
    \small
    \bibliographystyle{ieeenat_fullname}
    \bibliography{main}

@article{pavlakos2023reconstructing,
  title={Reconstructing Hands in 3D with Transformers},
  author={Pavlakos, Georgios and Shan, Dandan and Radosavovic, Ilija and Kanazawa, Angjoo and Fouhey, David and Malik, Jitendra},
  journal={arXiv preprint arXiv:2312.05251},
  year={2023}
}

@inproceedings{wu2023multiview,
  title={Multiview compressive coding for 3D reconstruction},
  author={Wu, Chao-Yuan and Johnson, Justin and Malik, Jitendra and Feichtenhofer, Christoph and Gkioxari, Georgia},
  booktitle={Proceedings of the IEEE/CVF Conference on Computer Vision and Pattern Recognition},
  pages={9065--9075},
  year={2023}
}

@inproceedings{reizenstein2021common,
  title={Common objects in 3d: Large-scale learning and evaluation of real-life 3d category reconstruction},
  author={Reizenstein, Jeremy and Shapovalov, Roman and Henzler, Philipp and Sbordone, Luca and Labatut, Patrick and Novotny, David},
  booktitle={Proceedings of the IEEE/CVF International Conference on Computer Vision},
  pages={10901--10911},
  year={2021}
}

@article{patel2022learning,
  title={Learning to imitate object interactions from internet videos},
  author={Patel, Austin and Wang, Andrew and Radosavovic, Ilija and Malik, Jitendra},
  journal={arXiv preprint arXiv:2211.13225},
  year={2022}
}

@inproceedings{cao2021reconstructing,
  title={Reconstructing hand-object interactions in the wild},
  author={Cao, Zhe and Radosavovic, Ilija and Kanazawa, Angjoo and Malik, Jitendra},
  booktitle={Proceedings of the IEEE/CVF International Conference on Computer Vision},
  pages={12417--12426},
  year={2021}
}

@article{choi2023handnerf,
  title={HandNeRF: Learning to Reconstruct Hand-Object Interaction Scene from a Single RGB Image},
  author={Choi, Hongsuk and Chavan-Dafle, Nikhil and Yuan, Jiacheng and Isler, Volkan and Park, Hyunsoo},
  journal={arXiv preprint arXiv:2309.07891},
  year={2023}
}

@inproceedings{liu2022hoi4d,
  title={HOI4D: A 4D egocentric dataset for category-level human-object interaction},
  author={Liu, Yunze and Liu, Yun and Jiang, Che and Lyu, Kangbo and Wan, Weikang and Shen, Hao and Liang, Boqiang and Fu, Zhoujie and Wang, He and Yi, Li},
  booktitle={Proceedings of the IEEE/CVF Conference on Computer Vision and Pattern Recognition},
  pages={21013--21022},
  year={2022}
}

@inproceedings{chao2021dexycb,
  title={DexYCB: A benchmark for capturing hand grasping of objects},
  author={Chao, Yu-Wei and Yang, Wei and Xiang, Yu and Molchanov, Pavlo and Handa, Ankur and Tremblay, Jonathan and Narang, Yashraj S and Van Wyk, Karl and Iqbal, Umar and Birchfield, Stan and others},
  booktitle={Proceedings of the IEEE/CVF Conference on Computer Vision and Pattern Recognition},
  pages={9044--9053},
  year={2021}
}

@INPROCEEDINGS{hampali2020honnotate,
	      title={HOnnotate: A method for 3D Annotation of Hand and Object Poses},
          author={Shreyas Hampali and Mahdi Rad and Markus Oberweger and Vincent Lepetit},
          booktitle = {CVPR},
      year = {2020}
         }

@misc{gpt4v,
    key = {OpenAI. Gpt-4v(ision) system card.},
    title = {OpenAI. Gpt-4v(ision) system card.},
    year = {2023}
}

@misc{luma,
    key = {Luma AI. Capture 3D. https://lumalabs.ai.},
    title = {Luma AI. Capture 3D. https://lumalabs.ai.},
    year = {2023}
}

@inproceedings{huang2022reconstructing,
  title={Reconstructing hand-held objects from monocular video},
  author={Huang, Di and Ji, Xiaopeng and He, Xingyi and Sun, Jiaming and He, Tong and Shuai, Qing and Ouyang, Wanli and Zhou, Xiaowei},
  booktitle={SIGGRAPH Asia 2022 Conference Papers},
  pages={1--9},
  year={2022}
}

@inproceedings{chen2023gsdf,
  title={gSDF: Geometry-Driven Signed Distance Functions for 3D Hand-Object Reconstruction},
  author={Chen, Zerui and Chen, Shizhe and Schmid, Cordelia and Laptev, Ivan},
  booktitle={Proceedings of the IEEE/CVF Conference on Computer Vision and Pattern Recognition},
  pages={12890--12900},
  year={2023}
}

@inproceedings{hampali2023hand,
  title={In-Hand 3D Object Scanning from an RGB Sequence},
  author={Hampali, Shreyas and Hodan, Tomas and Tran, Luan and Ma, Lingni and Keskin, Cem and Lepetit, Vincent},
  booktitle={Proceedings of the IEEE/CVF Conference on Computer Vision and Pattern Recognition},
  pages={17079--17088},
  year={2023}
}

@inproceedings{xie2022chore,
  title={Chore: Contact, human and object reconstruction from a single rgb image},
  author={Xie, Xianghui and Bhatnagar, Bharat Lal and Pons-Moll, Gerard},
  booktitle={European Conference on Computer Vision},
  pages={125--145},
  year={2022},
  organization={Springer}
}

@inproceedings{qu2023novel,
  title={Novel-view Synthesis and Pose Estimation for Hand-Object Interaction from Sparse Views},
  author={Qu, Wentian and Cui, Zhaopeng and Zhang, Yinda and Meng, Chenyu and Ma, Cuixia and Deng, Xiaoming and Wang, Hongan},
  booktitle={Proceedings of the IEEE/CVF International Conference on Computer Vision},
  pages={15100--15111},
  year={2023}
}

@inproceedings{ye2023diffusion,
  title={Diffusion-Guided Reconstruction of Everyday Hand-Object Interaction Clips},
  author={Ye, Yufei and Hebbar, Poorvi and Gupta, Abhinav and Tulsiani, Shubham},
  booktitle={Proceedings of the IEEE/CVF International Conference on Computer Vision},
  pages={19717--19728},
  year={2023}
}

@article{zhang2023moho,
  title={MOHO: Learning Single-view Hand-held Object Reconstruction with Multi-view Occlusion-Aware Supervision},
  author={Zhang, Chenyangguang and Jiao, Guanlong and Di, Yan and Huang, Ziqin and Wang, Gu and Zhang, Ruida and Fu, Bowen and Tombari, Federico and Ji, Xiangyang},
  journal={arXiv preprint arXiv:2310.11696},
  year={2023}
}

@article{fu2023hacd,
  title={HACD: Hand-Aware Conditional Diffusion for Monocular Hand-Held Object Reconstruction},
  author={Fu, Bowen and Di, Yan and Zhang, Chenyangguang and Wang, Gu and Huang, Ziqin and Leng, Zhiying and Manhardt, Fabian and Ji, Xiangyang and Tombari, Federico},
  journal={arXiv preprint arXiv:2311.14189},
  year={2023}
}

@article{romero2022embodied,
  title={Embodied hands: Modeling and capturing hands and bodies together},
  author={Romero, Javier and Tzionas, Dimitrios and Black, Michael J},
  journal={arXiv preprint arXiv:2201.02610},
  year={2022}
}

@inproceedings{tse2022collaborative,
  title={Collaborative learning for hand and object reconstruction with attention-guided graph convolution},
  author={Tse, Tze Ho Elden and Kim, Kwang In and Leonardis, Ales and Chang, Hyung Jin},
  booktitle={Proceedings of the IEEE/CVF Conference on Computer Vision and Pattern Recognition},
  pages={1664--1674},
  year={2022}
}

@inproceedings{yang2022artiboost,
  title={ArtiBoost: Boosting articulated 3d hand-object pose estimation via online exploration and synthesis},
  author={Yang, Lixin and Li, Kailin and Zhan, Xinyu and Lv, Jun and Xu, Wenqiang and Li, Jiefeng and Lu, Cewu},
  booktitle={Proceedings of the IEEE/CVF Conference on Computer Vision and Pattern Recognition},
  pages={2750--2760},
  year={2022}
}

@inproceedings{hasson2019learning,
  title={Learning joint reconstruction of hands and manipulated objects},
  author={Hasson, Yana and Varol, Gul and Tzionas, Dimitrios and Kalevatykh, Igor and Black, Michael J and Laptev, Ivan and Schmid, Cordelia},
  booktitle={Proceedings of the IEEE/CVF conference on computer vision and pattern recognition},
  pages={11807--11816},
  year={2019}
}

@inproceedings{hasson2020leveraging,
  title={Leveraging photometric consistency over time for sparsely supervised hand-object reconstruction},
  author={Hasson, Yana and Tekin, Bugra and Bogo, Federica and Laptev, Ivan and Pollefeys, Marc and Schmid, Cordelia},
  booktitle={Proceedings of the IEEE/CVF conference on computer vision and pattern recognition},
  pages={571--580},
  year={2020}
}

@inproceedings{bhatnagar2022behave,
  title={Behave: Dataset and method for tracking human object interactions},
  author={Bhatnagar, Bharat Lal and Xie, Xianghui and Petrov, Ilya A and Sminchisescu, Cristian and Theobalt, Christian and Pons-Moll, Gerard},
  booktitle={Proceedings of the IEEE/CVF Conference on Computer Vision and Pattern Recognition},
  pages={15935--15946},
  year={2022}
}

@inproceedings{fan2023arctic,
  title={ARCTIC: A Dataset for Dexterous Bimanual Hand-Object Manipulation},
  author={Fan, Zicong and Taheri, Omid and Tzionas, Dimitrios and Kocabas, Muhammed and Kaufmann, Manuel and Black, Michael J and Hilliges, Otmar},
  booktitle={Proceedings of the IEEE/CVF Conference on Computer Vision and Pattern Recognition},
  pages={12943--12954},
  year={2023}
}

@inproceedings{hasson2021towards,
  title={Towards unconstrained joint hand-object reconstruction from rgb videos},
  author={Hasson, Yana and Varol, G{\"u}l and Schmid, Cordelia and Laptev, Ivan},
  booktitle={2021 International Conference on 3D Vision (3DV)},
  pages={659--668},
  year={2021},
  organization={IEEE}
}

@inproceedings{tekin2019h+,
  title={H+ o: Unified egocentric recognition of 3d hand-object poses and interactions},
  author={Tekin, Bugra and Bogo, Federica and Pollefeys, Marc},
  booktitle={Proceedings of the IEEE/CVF conference on computer vision and pattern recognition},
  pages={4511--4520},
  year={2019}
}

@article{tzionas2016capturing,
  title={Capturing hands in action using discriminative salient points and physics simulation},
  author={Tzionas, Dimitrios and Ballan, Luca and Srikantha, Abhilash and Aponte, Pablo and Pollefeys, Marc and Gall, Juergen},
  journal={International Journal of Computer Vision},
  volume={118},
  pages={172--193},
  year={2016},
  publisher={Springer}
}

@inproceedings{li2022interacting,
  title={Interacting attention graph for single image two-hand reconstruction},
  author={Li, Mengcheng and An, Liang and Zhang, Hongwen and Wu, Lianpeng and Chen, Feng and Yu, Tao and Liu, Yebin},
  booktitle={Proceedings of the IEEE/CVF Conference on Computer Vision and Pattern Recognition},
  pages={2761--2770},
  year={2022}
}

@inproceedings{meng20223d,
  title={3d interacting hand pose estimation by hand de-occlusion and removal},
  author={Meng, Hao and Jin, Sheng and Liu, Wentao and Qian, Chen and Lin, Mengxiang and Ouyang, Wanli and Luo, Ping},
  booktitle={European Conference on Computer Vision},
  pages={380--397},
  year={2022},
  organization={Springer}
}

@inproceedings{moon2023bringing,
  title={Bringing Inputs to Shared Domains for 3D Interacting Hands Recovery in the Wild},
  author={Moon, Gyeongsik},
  booktitle={Proceedings of the IEEE/CVF Conference on Computer Vision and Pattern Recognition},
  pages={17028--17037},
  year={2023}
}

@inproceedings{wang2023memahand,
  title={MeMaHand: Exploiting Mesh-Mano Interaction for Single Image Two-Hand Reconstruction},
  author={Wang, Congyi and Zhu, Feida and Wen, Shilei},
  booktitle={Proceedings of the IEEE/CVF Conference on Computer Vision and Pattern Recognition},
  pages={564--573},
  year={2023}
}

@inproceedings{yu2023acr,
  title={ACR: Attention Collaboration-based Regressor for Arbitrary Two-Hand Reconstruction},
  author={Yu, Zhengdi and Huang, Shaoli and Fang, Chen and Breckon, Toby P and Wang, Jue},
  booktitle={Proceedings of the IEEE/CVF Conference on Computer Vision and Pattern Recognition},
  pages={12955--12964},
  year={2023}
}

@inproceedings{zuo2023reconstructing,
  title={Reconstructing interacting hands with interaction prior from monocular images},
  author={Zuo, Binghui and Zhao, Zimeng and Sun, Wenqian and Xie, Wei and Xue, Zhou and Wang, Yangang},
  booktitle={Proceedings of the IEEE/CVF International Conference on Computer Vision},
  pages={9054--9064},
  year={2023}
}

@inproceedings{park2022handoccnet,
  title={Handoccnet: Occlusion-robust 3d hand mesh estimation network},
  author={Park, JoonKyu and Oh, Yeonguk and Moon, Gyeongsik and Choi, Hongsuk and Lee, Kyoung Mu},
  booktitle={Proceedings of the IEEE/CVF Conference on Computer Vision and Pattern Recognition},
  pages={1496--1505},
  year={2022}
}

@inproceedings{gkioxari2019mesh,
  title={Mesh r-cnn},
  author={Gkioxari, Georgia and Malik, Jitendra and Johnson, Justin},
  booktitle={Proceedings of the IEEE/CVF international conference on computer vision},
  pages={9785--9795},
  year={2019}
}

@book{hartley2003multiple,
  title={Multiple view geometry in computer vision},
  author={Hartley, Richard and Zisserman, Andrew},
  year={2003},
  publisher={Cambridge university press}
}

@article{scharstein2002taxonomy,
  title={A taxonomy and evaluation of dense two-frame stereo correspondence algorithms},
  author={Scharstein, Daniel and Szeliski, Richard},
  journal={International journal of computer vision},
  volume={47},
  pages={7--42},
  year={2002},
  publisher={Springer}
}

@book{szeliski2022computer,
  title={Computer vision: algorithms and applications},
  author={Szeliski, Richard},
  year={2022},
  publisher={Springer Nature}
}

@article{tomasi1992shape,
  title={Shape and motion from image streams under orthography: a factorization method},
  author={Tomasi, Carlo and Kanade, Takeo},
  journal={International journal of computer vision},
  volume={9},
  pages={137--154},
  year={1992},
  publisher={Springer}
}

@article{mur2015orb,
  title={ORB-SLAM: a versatile and accurate monocular SLAM system},
  author={Mur-Artal, Raul and Montiel, Jose Maria Martinez and Tardos, Juan D},
  journal={IEEE transactions on robotics},
  volume={31},
  number={5},
  pages={1147--1163},
  year={2015},
  publisher={IEEE}
}

@article{castellanos1999spmap,
  title={The SPmap: A probabilistic framework for simultaneous localization and map building},
  author={Castellanos, Jose A and Montiel, Jos{\'e} MM and Neira, Jos{\'e} and Tard{\'o}s, Juan D},
  journal={IEEE Transactions on robotics and Automation},
  volume={15},
  number={5},
  pages={948--952},
  year={1999},
  publisher={IEEE}
}

@article{taketomi2017visual,
  title={Visual SLAM algorithms: A survey from 2010 to 2016},
  author={Taketomi, Takafumi and Uchiyama, Hideaki and Ikeda, Sei},
  journal={IPSJ Transactions on Computer Vision and Applications},
  volume={9},
  number={1},
  pages={1--11},
  year={2017},
  publisher={Springer}
}

@article{mildenhall2021nerf,
  title={Nerf: Representing scenes as neural radiance fields for view synthesis},
  author={Mildenhall, Ben and Srinivasan, Pratul P and Tancik, Matthew and Barron, Jonathan T and Ramamoorthi, Ravi and Ng, Ren},
  journal={Communications of the ACM},
  volume={65},
  number={1},
  pages={99--106},
  year={2021},
  publisher={ACM New York, NY, USA}
}

@inproceedings{park2019deepsdf,
  title={Deepsdf: Learning continuous signed distance functions for shape representation},
  author={Park, Jeong Joon and Florence, Peter and Straub, Julian and Newcombe, Richard and Lovegrove, Steven},
  booktitle={Proceedings of the IEEE/CVF conference on computer vision and pattern recognition},
  pages={165--174},
  year={2019}
}

@article{wang2021neus,
  title={Neus: Learning neural implicit surfaces by volume rendering for multi-view reconstruction},
  author={Wang, Peng and Liu, Lingjie and Liu, Yuan and Theobalt, Christian and Komura, Taku and Wang, Wenping},
  journal={arXiv preprint arXiv:2106.10689},
  year={2021}
}

@article{muller2022instant,
  title={Instant neural graphics primitives with a multiresolution hash encoding},
  author={M{\"u}ller, Thomas and Evans, Alex and Schied, Christoph and Keller, Alexander},
  journal={ACM Transactions on Graphics (ToG)},
  volume={41},
  number={4},
  pages={1--15},
  year={2022},
  publisher={ACM New York, NY, USA}
}

@inproceedings{tewari2020state,
  title={State of the art on neural rendering},
  author={Tewari, Ayush and Fried, Ohad and Thies, Justus and Sitzmann, Vincent and Lombardi, Stephen and Sunkavalli, Kalyan and Martin-Brualla, Ricardo and Simon, Tomas and Saragih, Jason and Nie{\ss}ner, Matthias and others},
  booktitle={Computer Graphics Forum},
  volume={39},
  pages={701--727},
  year={2020},
  organization={Wiley Online Library}
}

@inproceedings{deitke2023objaverse,
  title={Objaverse: A universe of annotated 3d objects},
  author={Deitke, Matt and Schwenk, Dustin and Salvador, Jordi and Weihs, Luca and Michel, Oscar and VanderBilt, Eli and Schmidt, Ludwig and Ehsani, Kiana and Kembhavi, Aniruddha and Farhadi, Ali},
  booktitle={Proceedings of the IEEE/CVF Conference on Computer Vision and Pattern Recognition},
  pages={13142--13153},
  year={2023}
}

@article{deitke2024objaverse,
  title={Objaverse-xl: A universe of 10m+ 3d objects},
  author={Deitke, Matt and Liu, Ruoshi and Wallingford, Matthew and Ngo, Huong and Michel, Oscar and Kusupati, Aditya and Fan, Alan and Laforte, Christian and Voleti, Vikram and Gadre, Samir Yitzhak and others},
  journal={Advances in Neural Information Processing Systems},
  volume={36},
  year={2024}
}

@inproceedings{wang2021multi,
  title={Multi-view 3d reconstruction with transformers},
  author={Wang, Dan and Cui, Xinrui and Chen, Xun and Zou, Zhengxia and Shi, Tianyang and Salcudean, Septimiu and Wang, Z Jane and Ward, Rabab},
  booktitle={Proceedings of the IEEE/CVF International Conference on Computer Vision},
  pages={5722--5731},
  year={2021}
}

@inproceedings{xie2019pix2vox,
  title={Pix2vox: Context-aware 3d reconstruction from single and multi-view images},
  author={Xie, Haozhe and Yao, Hongxun and Sun, Xiaoshuai and Zhou, Shangchen and Zhang, Shengping},
  booktitle={Proceedings of the IEEE/CVF international conference on computer vision},
  pages={2690--2698},
  year={2019}
}

@inproceedings{liu2023zero,
  title={Zero-1-to-3: Zero-shot one image to 3d object},
  author={Liu, Ruoshi and Wu, Rundi and Van Hoorick, Basile and Tokmakov, Pavel and Zakharov, Sergey and Vondrick, Carl},
  booktitle={Proceedings of the IEEE/CVF International Conference on Computer Vision},
  pages={9298--9309},
  year={2023}
}

@article{liu2024one,
  title={One-2-3-45: Any single image to 3d mesh in 45 seconds without per-shape optimization},
  author={Liu, Minghua and Xu, Chao and Jin, Haian and Chen, Linghao and Varma T, Mukund and Xu, Zexiang and Su, Hao},
  journal={Advances in Neural Information Processing Systems},
  volume={36},
  year={2024}
}

@inproceedings{qi2017pointnet,
  title={Pointnet: Deep learning on point sets for 3d classification and segmentation},
  author={Qi, Charles R and Su, Hao and Mo, Kaichun and Guibas, Leonidas J},
  booktitle={Proceedings of the IEEE conference on computer vision and pattern recognition},
  pages={652--660},
  year={2017}
}

@article{qi2017pointnet++,
  title={Pointnet++: Deep hierarchical feature learning on point sets in a metric space},
  author={Qi, Charles Ruizhongtai and Yi, Li and Su, Hao and Guibas, Leonidas J},
  journal={Advances in neural information processing systems},
  volume={30},
  year={2017}
}

@inproceedings{wen2019pixel2mesh++,
  title={Pixel2mesh++: Multi-view 3d mesh generation via deformation},
  author={Wen, Chao and Zhang, Yinda and Li, Zhuwen and Fu, Yanwei},
  booktitle={Proceedings of the IEEE/CVF international conference on computer vision},
  pages={1042--1051},
  year={2019}
}

@inproceedings{worchel2022multi,
  title={Multi-view mesh reconstruction with neural deferred shading},
  author={Worchel, Markus and Diaz, Rodrigo and Hu, Weiwen and Schreer, Oliver and Feldmann, Ingo and Eisert, Peter},
  booktitle={Proceedings of the IEEE/CVF Conference on Computer Vision and Pattern Recognition},
  pages={6187--6197},
  year={2022}
}

@inproceedings{reizenstein21co3d,
	Author = {Reizenstein, Jeremy and Shapovalov, Roman and Henzler, Philipp and Sbordone, Luca and Labatut, Patrick and Novotny, David},
	Booktitle = {International Conference on Computer Vision},
	Title = {Common Objects in 3D: Large-Scale Learning and Evaluation of Real-life 3D Category Reconstruction},
	Year = {2021},
}

@inproceedings{lin2014microsoft,
  title={Microsoft coco: Common objects in context},
  author={Lin, Tsung-Yi and Maire, Michael and Belongie, Serge and Hays, James and Perona, Pietro and Ramanan, Deva and Doll{\'a}r, Piotr and Zitnick, C Lawrence},
  booktitle={Computer Vision--ECCV 2014: 13th European Conference, Zurich, Switzerland, September 6-12, 2014, Proceedings, Part V 13},
  pages={740--755},
  year={2014},
  organization={Springer}
}

@article{long2023wonder3d,
  title={Wonder3d: Single image to 3d using cross-domain diffusion},
  author={Long, Xiaoxiao and Guo, Yuan-Chen and Lin, Cheng and Liu, Yuan and Dou, Zhiyang and Liu, Lingjie and Ma, Yuexin and Zhang, Song-Hai and Habermann, Marc and Theobalt, Christian and others},
  journal={arXiv preprint arXiv:2310.15008},
  year={2023}
}

@article{hong2023lrm,
  title={Lrm: Large reconstruction model for single image to 3d},
  author={Hong, Yicong and Zhang, Kai and Gu, Jiuxiang and Bi, Sai and Zhou, Yang and Liu, Difan and Liu, Feng and Sunkavalli, Kalyan and Bui, Trung and Tan, Hao},
  journal={arXiv preprint arXiv:2311.04400},
  year={2023}
}

@article{ravi2020accelerating,
  title={Accelerating 3d deep learning with pytorch3d},
  author={Ravi, Nikhila and Reizenstein, Jeremy and Novotny, David and Gordon, Taylor and Lo, Wan-Yen and Johnson, Justin and Gkioxari, Georgia},
  journal={arXiv preprint arXiv:2007.08501},
  year={2020}
}

@inproceedings{loper2014opendr,
  title={OpenDR: An approximate differentiable renderer},
  author={Loper, Matthew M and Black, Michael J},
  booktitle={Computer Vision--ECCV 2014: 13th European Conference, Zurich, Switzerland, September 6-12, 2014, Proceedings, Part VII 13},
  pages={154--169},
  year={2014},
  organization={Springer}
}

@inproceedings{wen2023bundlesdf,
  title={BundleSDF: Neural 6-DoF Tracking and 3D Reconstruction of Unknown Objects},
  author={Wen, Bowen and Tremblay, Jonathan and Blukis, Valts and Tyree, Stephen and M{\"u}ller, Thomas and Evans, Alex and Fox, Dieter and Kautz, Jan and Birchfield, Stan},
  booktitle={Proceedings of the IEEE/CVF Conference on Computer Vision and Pattern Recognition},
  pages={606--617},
  year={2023}
}

@inproceedings{ye2022s,
  title={What's in your hands? 3D Reconstruction of Generic Objects in Hands},
  author={Ye, Yufei and Gupta, Abhinav and Tulsiani, Shubham},
  booktitle={Proceedings of the IEEE/CVF Conference on Computer Vision and Pattern Recognition},
  pages={3895--3905},
  year={2022}
}

@inproceedings{kwon2021h2o,
  title={H2o: Two hands manipulating objects for first person interaction recognition},
  author={Kwon, Taein and Tekin, Bugra and St{\"u}hmer, Jan and Bogo, Federica and Pollefeys, Marc},
  booktitle={Proceedings of the IEEE/CVF International Conference on Computer Vision},
  pages={10138--10148},
  year={2021}
}

@inproceedings{damen2018scaling,
  title={Scaling egocentric vision: The epic-kitchens dataset},
  author={Damen, Dima and Doughty, Hazel and Farinella, Giovanni Maria and Fidler, Sanja and Furnari, Antonino and Kazakos, Evangelos and Moltisanti, Davide and Munro, Jonathan and Perrett, Toby and Price, Will and others},
  booktitle={Proceedings of the European conference on computer vision (ECCV)},
  pages={720--736},
  year={2018}
}

@inproceedings{grauman2022ego4d,
  title={Ego4d: Around the world in 3,000 hours of egocentric video},
  author={Grauman, Kristen and Westbury, Andrew and Byrne, Eugene and Chavis, Zachary and Furnari, Antonino and Girdhar, Rohit and Hamburger, Jackson and Jiang, Hao and Liu, Miao and Liu, Xingyu and others},
  booktitle={Proceedings of the IEEE/CVF Conference on Computer Vision and Pattern Recognition},
  pages={18995--19012},
  year={2022}
}

@article{grauman2023ego,
  title={Ego-exo4d: Understanding skilled human activity from first-and third-person perspectives},
  author={Grauman, Kristen and Westbury, Andrew and Torresani, Lorenzo and Kitani, Kris and Malik, Jitendra and Afouras, Triantafyllos and Ashutosh, Kumar and Baiyya, Vijay and Bansal, Siddhant and Boote, Bikram and others},
  journal={arXiv preprint arXiv:2311.18259},
  year={2023}
}

@inproceedings{chen2022alignsdf,
  title={Alignsdf: Pose-aligned signed distance fields for hand-object reconstruction},
  author={Chen, Zerui and Hasson, Yana and Schmid, Cordelia and Laptev, Ivan},
  booktitle={European Conference on Computer Vision},
  pages={231--248},
  year={2022},
  organization={Springer}
}

@inproceedings{shan2020understanding,
  title={Understanding human hands in contact at internet scale},
  author={Shan, Dandan and Geng, Jiaqi and Shu, Michelle and Fouhey, David F},
  booktitle={Proceedings of the IEEE/CVF conference on computer vision and pattern recognition},
  pages={9869--9878},
  year={2020}
}

@article{kirillov2023segment,
  title={Segment anything},
  author={Kirillov, Alexander and Mintun, Eric and Ravi, Nikhila and Mao, Hanzi and Rolland, Chloe and Gustafson, Laura and Xiao, Tete and Whitehead, Spencer and Berg, Alexander C and Lo, Wan-Yen and others},
  journal={arXiv preprint arXiv:2304.02643},
  year={2023}
}

@article{vaswani2017attention,
  title={Attention is all you need},
  author={Vaswani, Ashish and Shazeer, Noam and Parmar, Niki and Uszkoreit, Jakob and Jones, Llion and Gomez, Aidan N and Kaiser, {\L}ukasz and Polosukhin, Illia},
  journal={Advances in neural information processing systems},
  volume={30},
  year={2017}
}

@article{dosovitskiy2020image,
  title={An image is worth 16x16 words: Transformers for image recognition at scale},
  author={Dosovitskiy, Alexey and Beyer, Lucas and Kolesnikov, Alexander and Weissenborn, Dirk and Zhai, Xiaohua and Unterthiner, Thomas and Dehghani, Mostafa and Minderer, Matthias and Heigold, Georg and Gelly, Sylvain and others},
  journal={arXiv preprint arXiv:2010.11929},
  year={2020}
}

@inproceedings{he2022masked,
  title={Masked autoencoders are scalable vision learners},
  author={He, Kaiming and Chen, Xinlei and Xie, Saining and Li, Yanghao and Doll{\'a}r, Piotr and Girshick, Ross},
  booktitle={Proceedings of the IEEE/CVF conference on computer vision and pattern recognition},
  pages={16000--16009},
  year={2022}
}

@article{lewis2020retrieval,
  title={Retrieval-augmented generation for knowledge-intensive nlp tasks},
  author={Lewis, Patrick and Perez, Ethan and Piktus, Aleksandra and Petroni, Fabio and Karpukhin, Vladimir and Goyal, Naman and K{\"u}ttler, Heinrich and Lewis, Mike and Yih, Wen-tau and Rockt{\"a}schel, Tim and others},
  journal={Advances in Neural Information Processing Systems},
  volume={33},
  pages={9459--9474},
  year={2020}
}

@article{li2023object,
  title={Object motion guided human motion synthesis},
  author={Li, Jiaman and Wu, Jiajun and Liu, C Karen},
  journal={ACM Transactions on Graphics (TOG)},
  volume={42},
  number={6},
  pages={1--11},
  year={2023},
  publisher={ACM New York, NY, USA}
}

@article{poole2022dreamfusion,
  title={Dreamfusion: Text-to-3d using 2d diffusion},
  author={Poole, Ben and Jain, Ajay and Barron, Jonathan T and Mildenhall, Ben},
  journal={arXiv preprint arXiv:2209.14988},
  year={2022}
}

@article{rong2020frankmocap,
  title={FrankMocap: Fast Monocular 3D Hand and Body Motion Capture by Regression and Integration},
  author={Rong, Yu and Shiratori, Takaaki and Joo, Hanbyul},
  journal={arXiv:2008.08324},
  year={2020}
}

@InProceedings{Karunratanakul_2023_CVPR,
    author    = {Karunratanakul, Korrawe and Prokudin, Sergey and Hilliges, Otmar and Tang, Siyu},
    title     = {HARP: Personalized Hand Reconstruction From a Monocular RGB Video},
    booktitle = {Proceedings of the IEEE/CVF Conference on Computer Vision and Pattern Recognition (CVPR)},
    month     = {June},
    year      = {2023},
    pages     = {12802-12813}
}

@InProceedings{Yu_2023_CVPR,
    author    = {Yu, Zhengdi and Huang, Shaoli and Fang, Chen and Breckon, Toby P. and Wang, Jue},
    title     = {ACR: Attention Collaboration-Based Regressor for Arbitrary Two-Hand Reconstruction},
    booktitle = {Proceedings of the IEEE/CVF Conference on Computer Vision and Pattern Recognition (CVPR)},
    month     = {June},
    year      = {2023},
    pages     = {12955-12964}
}

@InProceedings{Wang_2023_CVPR,
    author    = {Wang, Congyi and Zhu, Feida and Wen, Shilei},
    title     = {MeMaHand: Exploiting Mesh-Mano Interaction for Single Image Two-Hand Reconstruction},
    booktitle = {Proceedings of the IEEE/CVF Conference on Computer Vision and Pattern Recognition (CVPR)},
    month     = {June},
    year      = {2023},
    pages     = {564-573}
}

@InProceedings{Wen_2023_CVPR,
    author    = {Wen, Yilin and Pan, Hao and Yang, Lei and Pan, Jia and Komura, Taku and Wang, Wenping},
    title     = {Hierarchical Temporal Transformer for 3D Hand Pose Estimation and Action Recognition From Egocentric RGB Videos},
    booktitle = {Proceedings of the IEEE/CVF Conference on Computer Vision and Pattern Recognition (CVPR)},
    month     = {June},
    year      = {2023},
    pages     = {21243-21253}
}

@InProceedings{Xu_2023_CVPR,
    author    = {Xu, Hao and Wang, Tianyu and Tang, Xiao and Fu, Chi-Wing},
    title     = {H2ONet: Hand-Occlusion-and-Orientation-Aware Network for Real-Time 3D Hand Mesh Reconstruction},
    booktitle = {Proceedings of the IEEE/CVF Conference on Computer Vision and Pattern Recognition (CVPR)},
    month     = {June},
    year      = {2023},
    pages     = {17048-17058}
}

@article{Ranftl2021,
	author    = {Ren\'{e} Ranftl and Alexey Bochkovskiy and Vladlen Koltun},
	title     = {Vision Transformers for Dense Prediction},
	journal   = {ArXiv preprint},
	year      = {2021},
}

@article{forney1973viterbi,
  title={The viterbi algorithm},
  author={Forney, G David},
  journal={Proceedings of the IEEE},
  volume={61},
  number={3},
  pages={268--278},
  year={1973},
  publisher={Ieee}
}

@book{jurafsky2000speech,
  title={Speech \& language processing},
  author={Jurafsky, Dan},
  year={2000},
  publisher={Pearson Education India}
}

@article{viterbi1967error,
  title={Error bounds for convolutional codes and an asymptotically optimum decoding algorithm},
  author={Viterbi, Andrew},
  journal={IEEE transactions on Information Theory},
  volume={13},
  number={2},
  pages={260--269},
  year={1967},
  publisher={IEEE}
}

@article{oquab2023dinov2,
  title={Dinov2: Learning robust visual features without supervision},
  author={Oquab, Maxime and Darcet, Timoth{\'e}e and Moutakanni, Th{\'e}o and Vo, Huy and Szafraniec, Marc and Khalidov, Vasil and Fernandez, Pierre and Haziza, Daniel and Massa, Francisco and El-Nouby, Alaaeldin and others},
  journal={arXiv preprint arXiv:2304.07193},
  year={2023}
}

@article{ravi2024sam,
  title={Sam 2: Segment anything in images and videos},
  author={Ravi, Nikhila and Gabeur, Valentin and Hu, Yuan-Ting and Hu, Ronghang and Ryali, Chaitanya and Ma, Tengyu and Khedr, Haitham and R{\"a}dle, Roman and Rolland, Chloe and Gustafson, Laura and others},
  journal={arXiv preprint arXiv:2408.00714},
  year={2024}
}

@inproceedings{Prakash2024HOI,
    author = {Prakash, Aditya and Chang, Matthew and Jin, Matthew and Tu, Ruisen and Gupta, Saurabh},
    title = {3D Reconstruction of Objects in Hands without Real World 3D Supervision},
    booktitle = {European Conference on Computer Vision (ECCV)},
    year = {2024}
}

@inproceedings{prakash20243d,
  title={3d hand pose estimation in everyday egocentric images},
  author={Prakash, Aditya and Tu, Ruisen and Chang, Matthew and Gupta, Saurabh},
  booktitle={Proceedings of the European Conference on Computer Vision (ECCV)},
  year={2024}
}

@article{ye2024g,
  title={G-HOP: Generative Hand-Object Prior for Interaction Reconstruction and Grasp Synthesis},
  author={Ye, Yufei and Gupta, Abhinav and Kitani, Kris and Tulsiani, Shubham},
  journal={arXiv preprint arXiv:2404.12383},
  year={2024}
}

@article{zhang2023handypriors,
  title={HandyPriors: Physically Consistent Perception of Hand-Object Interactions with Differentiable Priors},
  author={Zhang, Shutong and Qiao, Yi-Ling and Zhu, Guanglei and Heiden, Eric and Turpin, Dylan and Liu, Jingzhou and Lin, Ming and Macklin, Miles and Garg, Animesh},
  journal={arXiv preprint arXiv:2311.16552},
  year={2023}
}

@inproceedings{kurz2017discretization,
  title={Discretization of SO (3) using recursive tesseract subdivision},
  author={Kurz, Gerhard and Pfaff, Florian and Hanebeck, Uwe D},
  booktitle={2017 IEEE International Conference on Multisensor Fusion and Integration for Intelligent Systems (MFI)},
  pages={49--55},
  year={2017},
  organization={IEEE}
}

@inproceedings{chen2024text,
  title={Text-to-3d using gaussian splatting},
  author={Chen, Zilong and Wang, Feng and Wang, Yikai and Liu, Huaping},
  booktitle={Proceedings of the IEEE/CVF Conference on Computer Vision and Pattern Recognition},
  pages={21401--21412},
  year={2024}
}

@inproceedings{long2024wonder3d,
  title={Wonder3d: Single image to 3d using cross-domain diffusion},
  author={Long, Xiaoxiao and Guo, Yuan-Chen and Lin, Cheng and Liu, Yuan and Dou, Zhiyang and Liu, Lingjie and Ma, Yuexin and Zhang, Song-Hai and Habermann, Marc and Theobalt, Christian and others},
  booktitle={Proceedings of the IEEE/CVF Conference on Computer Vision and Pattern Recognition},
  pages={9970--9980},
  year={2024}
}

@inproceedings{brahmbhatt2020contactpose,
  title={ContactPose: A dataset of grasps with object contact and hand pose},
  author={Brahmbhatt, Samarth and Tang, Chengcheng and Twigg, Christopher D and Kemp, Charles C and Hays, James},
  booktitle={Computer Vision--ECCV 2020: 16th European Conference, Glasgow, UK, August 23--28, 2020, Proceedings, Part XIII 16},
  pages={361--378},
  year={2020},
  organization={Springer}
}

@inproceedings{corona2020ganhand,
  title={Ganhand: Predicting human grasp affordances in multi-object scenes},
  author={Corona, Enric and Pumarola, Albert and Alenya, Guillem and Moreno-Noguer, Francesc and Rogez, Gr{\'e}gory},
  booktitle={Proceedings of the IEEE/CVF conference on computer vision and pattern recognition},
  pages={5031--5041},
  year={2020}
}

@inproceedings{taheri2020grab,
  title={GRAB: A dataset of whole-body human grasping of objects},
  author={Taheri, Omid and Ghorbani, Nima and Black, Michael J and Tzionas, Dimitrios},
  booktitle={Computer Vision--ECCV 2020: 16th European Conference, Glasgow, UK, August 23--28, 2020, Proceedings, Part IV 16},
  pages={581--600},
  year={2020},
  organization={Springer}
}

@inproceedings{yang2022oakink,
  title={Oakink: A large-scale knowledge repository for understanding hand-object interaction},
  author={Yang, Lixin and Li, Kailin and Zhan, Xinyu and Wu, Fei and Xu, Anran and Liu, Liu and Lu, Cewu},
  booktitle={Proceedings of the IEEE/CVF conference on computer vision and pattern recognition},
  pages={20953--20962},
  year={2022}
}

@inproceedings{potamias2025wilor,
  title={Wilor: End-to-end 3d hand localization and reconstruction in-the-wild},
  author={Potamias, Rolandos Alexandros and Zhang, Jinglei and Deng, Jiankang and Zafeiriou, Stefanos},
  booktitle={Proceedings of the Computer Vision and Pattern Recognition Conference},
  pages={12242--12254},
  year={2025}
}

@article{ye2025predicting,
  title={Predicting 4d hand trajectory from monocular videos},
  author={Ye, Yufei and Feng, Yao and Taheri, Omid and Feng, Haiwen and Tulsiani, Shubham and Black, Michael J},
  journal={arXiv preprint arXiv:2501.08329},
  year={2025}
}

@article{xu2024instantmesh,
  title={Instantmesh: Efficient 3d mesh generation from a single image with sparse-view large reconstruction models},
  author={Xu, Jiale and Cheng, Weihao and Gao, Yiming and Wang, Xintao and Gao, Shenghua and Shan, Ying},
  journal={arXiv preprint arXiv:2404.07191},
  year={2024}
}

@article{comanici2025gemini,
  title={Gemini 2.5: Pushing the frontier with advanced reasoning, multimodality, long context, and next generation agentic capabilities},
  author={Comanici, Gheorghe and Bieber, Eric and Schaekermann, Mike and Pasupat, Ice and Sachdeva, Noveen and Dhillon, Inderjit and Blistein, Marcel and Ram, Ori and Zhang, Dan and Rosen, Evan and others},
  journal={arXiv preprint arXiv:2507.06261},
  year={2025}
}

@article{singh2024hand,
  title={Hand-object interaction pretraining from videos},
  author={Singh, Himanshu Gaurav and Loquercio, Antonio and Sferrazza, Carmelo and Wu, Jane and Qi, Haozhi and Abbeel, Pieter and Malik, Jitendra},
  journal={arXiv preprint arXiv:2409.08273},
  year={2024}
}

@inproceedings{fu2025gigahands,
  title={Gigahands: A massive annotated dataset of bimanual hand activities},
  author={Fu, Rao and Zhang, Dingxi and Jiang, Alex and Fu, Wanjia and Funk, Austin and Ritchie, Daniel and Sridhar, Srinath},
  booktitle={Proceedings of the Computer Vision and Pattern Recognition Conference},
  pages={17461--17474},
  year={2025}
}

@article{hoque2025egodex,
  title={EgoDex: Learning Dexterous Manipulation from Large-Scale Egocentric Video},
  author={Hoque, Ryan and Huang, Peide and Yoon, David J and Sivapurapu, Mouli and Zhang, Jian},
  journal={arXiv preprint arXiv:2505.11709},
  year={2025}
}

@article{banerjee2024introducing,
  title={Introducing HOT3D: An egocentric dataset for 3D hand and object tracking},
  author={Banerjee, Prithviraj and Shkodrani, Sindi and Moulon, Pierre and Hampali, Shreyas and Zhang, Fan and Fountain, Jade and Miller, Edward and Basol, Selen and Newcombe, Richard and Wang, Robert and others},
  journal={arXiv preprint arXiv:2406.09598},
  year={2024}
}
}

\end{document}